\documentclass[10pt,twocolumn,letterpaper]{article}
\usepackage{authblk}

\usepackage{cvpr}              

\usepackage{bm}
\usepackage{booktabs}
\usepackage{tabularx}
\usepackage{float}
\usepackage[dvipsnames]{xcolor}
\usepackage{graphicx} \usepackage{multirow}
\usepackage{amsmath}
\usepackage{amssymb}
\usepackage[numbers,sort&compress]{natbib}
\usepackage{colortbl}
\usepackage{adjustbox}

\usepackage[accsupp]{axessibility}

\usepackage{tikz}
\usetikzlibrary{positioning}

\usepackage{pgfplots}
\pgfplotsset{compat=1.18}

\newcommand{\vect}[1]{\bm{#1}}
\newcommand{\mat}[1]{\bm{#1}}

\DeclareMathOperator{\Var}{Var}

\DeclareRobustCommand\onedot{\futurelet\@let@token\@onedot}
\def\onedot{\ifx\@let@token.\else.\null\fi\xspace}
\def\eg{\emph{e.g}\onedot}

\def\etal{\emph{et al}\onedot}

\newcolumntype{Y}{>{\centering\arraybackslash}X}

\definecolor{lightgrey}{rgb}{0.9, 0.9, .9}

\newcommand{\ours}{OpenSplat3D\xspace}

\newcommand{\PAR}[1]{\vskip4pt \noindent {\bf #1~}}
\newcommand{\PARbegin}[1]{\noindent {\bf #1~}}

\definecolor{cvprblue}{rgb}{0.21,0.49,0.74}
\usepackage[pagebackref,breaklinks,colorlinks,allcolors=cvprblue]{hyperref}

\def\paperID{21} \def\confName{CVPR}
\def\confYear{2025}

\title{OpenSplat3D: Open-Vocabulary 3D Instance \\ Segmentation using Gaussian Splatting}

\makeatletter
\renewcommand\AB@affilsepx{\qquad\protect\Affilfont}
\newcommand\email[2][]{\newaffiltrue\let\AB@blk@and\AB@pand
      \if\relax#1\relax\def\AB@note{\AB@thenote}\else\def\AB@note{\relax}\setcounter{Maxaffil}{0}\fi
      \begingroup
        \let\protect\@unexpandable@protect
        \def\thanks{\protect\thanks}\def\footnote{\protect\footnote}\@temptokena=\expandafter{\AB@authors}{\def\\{\protect\\\protect\Affilfont}\xdef\AB@temp{#2}}\xdef\AB@authors{\the\@temptokena\AB@las\AB@au@str
         \protect\\[\affilsep]\protect\Affilfont\AB@temp}\gdef\AB@las{}\gdef\AB@au@str{}{\def\\{, \ignorespaces}\xdef\AB@temp{#2}}\@temptokena=\expandafter{\AB@affillist}\xdef\AB@affillist{\the\@temptokena \AB@affilsep
          \AB@affilnote{}\protect\Affilfont\AB@temp}\endgroup
       \let\AB@affilsep\AB@affilsepx
}
\makeatother

\author[1]{Jens Piekenbrinck}
\author[1]{Christian Schmidt}
\author[1]{Alexander Hermans}
\author[2]{Narunas Vaskevicius}
\author[2]{Timm Linder}
\author[1]{Bastian Leibe}

\affil[1]{RWTH Aachen University}
\affil[2]{Robert Bosch GmbH}

\begin{document}
\maketitle
\begin{abstract}
3D Gaussian Splatting (3DGS) has emerged as a powerful representation for neural scene reconstruction, offering high-quality novel view synthesis while maintaining computational efficiency.
In this paper, we extend the capabilities of 3DGS beyond pure scene representation by introducing an approach for open-vocabulary 3D instance segmentation without requiring manual labeling, termed OpenSplat3D.
Our method leverages feature-splatting techniques to associate semantic information with individual Gaussians, enabling fine-grained scene understanding.
We incorporate Segment Anything Model instance masks with a contrastive loss formulation as guidance for the instance features to achieve accurate instance-level segmentation.
Furthermore, we utilize language embeddings of a vision-language model, allowing for flexible, text-driven instance identification.
This combination enables our system to identify and segment arbitrary objects in 3D scenes based on natural language descriptions.
We show results on LERF-mask and LERF-OVS as well as the full ScanNet++ validation set, demonstrating the effectiveness of our approach.
\end{abstract}
\vspace{-10pt}
\section{Introduction}
\label{sec:intro}

Understanding and interpreting complex 3D scenes is a crucial challenge in computer vision, with applications spanning from robotics and augmented reality to autonomous systems.
Traditional 3D instance segmentation methods rely heavily on manually labeled datasets, a process that is labor intensive and inherently limited to predefined object categories.
This limitation has raised interest in open-vocabulary 3D scene understanding, where models are expected to segment and identify arbitrary objects without being restricted by a fixed set of labels.
Additionally, most 3D instance segmentation methods rely on 3D point cloud datasets, which poses an additional constraint, as recording 3D data requires specialized recording setups \cite{yeshwanth2023scannet++}.
This constraint limits dataset scalability, restricting the diversity and size of available 3D datasets.

\begin{figure}
    \centering
\includegraphics[width=1\linewidth]{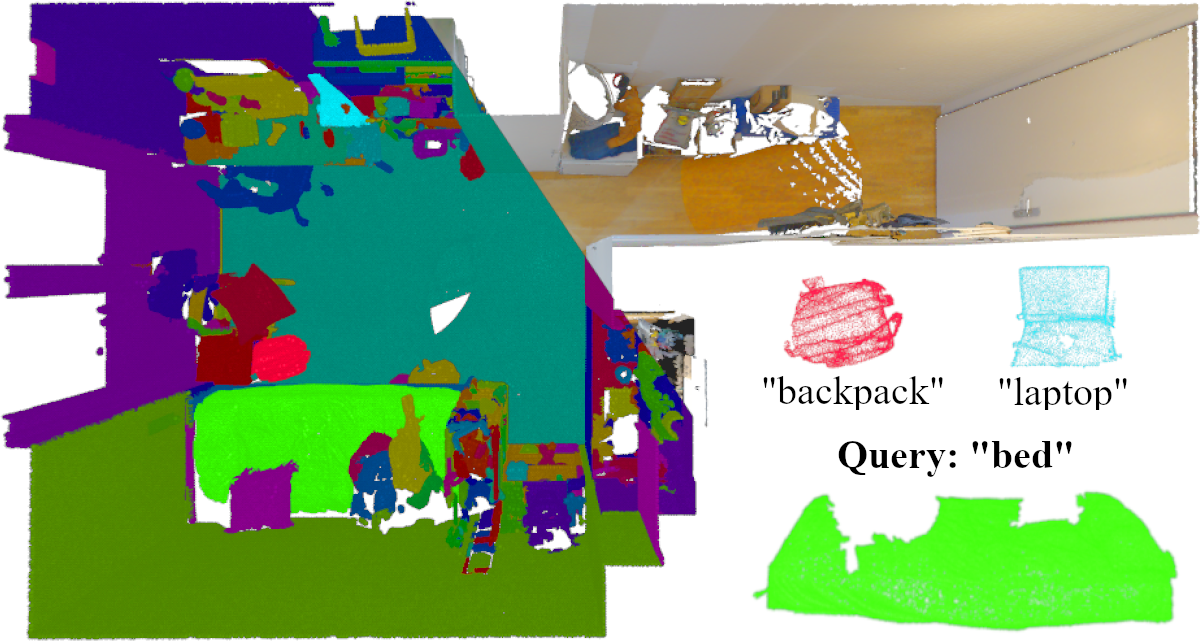}
\caption{OpenSplat3D jointly optimizes 3D geometry and 3D instance segmentation of a scene based on a sequence of 2D input images using 3D Gaussian Splatting.
    We project the resulting 3D segmentations to the 2D views and extract per-instance language features.
    This allows us to perform open-vocabulary 3D instance segmentation, here demonstrated on a room scene of ScanNet++.}
\label{fig:teaser}
\end{figure}

In the domain of 2D image understanding, the development of powerful foundation models \cite{radford2021clip,zhai2023siglip,kirillov2023sam,caron2021dino,oquab2023dinov2} has sparked a wave of research into open-vocabulary and language guided recognition.
However, the underlying foundation models are only available for 2D image data, and the required amount of annotated 3D data to train equivalent native 3D foundation models currently does not exist.
Previous work has therefore focused on projecting or embedding features from existing 2D foundation models into 3D point cloud representations \cite{yang2023sam3d,takmaz2023openmask3d,huang2025segment3d}.
While effective to some extent, 3D point clouds have notable limitations as their sparsity reduces feature coverage, they struggle with occlusion handling of surfaces and therefore a dense and differentiable 2D-to-3D mapping is not available.

In this work, we leverage multi-view data, and explore how 3D Gaussian Splatting (3DGS) \cite{kerbl2023splatting}, a state-of-the-art representation for neural scene reconstruction, can be extended beyond its original focus on novel view synthesis to support zero-shot open-vocabulary 3D instance segmentation.
By combining differentiable rendering with 2D semantic cues, we introduce a framework capable of assigning semantic and instance-level labels to individual 3D Gaussians without requiring any manual 3D annotations.

Our method integrates SAM~\cite{kirillov2023sam}, a 2D foundational segmentation model for instance mask generation, as well as language embeddings from vision-language models \cite{radford2021clip,zhai2023siglip,xu2023masqclip} for text-driven object identification.
These signals are projected into the 3D domain using differentiable rendering, allowing seamless alignment between 2D observations and the 3D Gaussian representation.

While earlier 3D Gaussian-based methods typically address \emph{either} instance-level reasoning \emph{or} language-based semantic retrieval in isolation \cite{qin2024langsplat,ye2024gaussiangrouping,gu2024egolifter,silva2024contrastive}, OpenGaussian \cite{wu2024opengaussian} very recently demonstrated that both tasks can be integrated into a joint approach. Similar to them, our concurrently developed method also unifies point-level instance segmentation and open-vocabulary semantics into a combined framework. This allows for an interesting quantitative comparison to recent \emph{point cloud-based} open-vocabulary instance segmentation methods like Segment3D \cite{huang2025segment3d}. 

Additionally, we introduce a novel variance regularization loss and a pixel-wise contrastive loss inspired by \cite{gu2024egolifter, wu2024opengaussian}, to ensure robust and consistent segmentation of different instances independent of the viewpoint.

We evaluate our approach on the challenging ScanNet++(v1)~\cite{yeshwanth2023scannet++} dataset following the class-agnostic approach of Segment3D~\cite{huang2025segment3d}, demonstrating that our method effectively segments instances in complex indoor scenes.
We also showcase our open-vocabulary segmentation capabilities on the LERF-mask~\cite{ye2024gaussiangrouping} and LERF-OVS~\cite{qin2024langsplat} dataset. Our approach significantly outperforms the recently introduced OpenGaussian and Segment3D methods.

\section{Related Work}
\label{sec:related_work}

\PARbegin{Foundation Models.}
Foundation models \cite{kirillov2023sam,radford2021clip,caron2021dino,oquab2023dinov2} have driven major advances in 2D image understanding.
These models, specifically designed for image-based tasks, exhibit remarkable zero-shot capabilities, allowing a wide range of applications without the need for task-specific fine-tuning.
Notably, CLIP~\cite{radford2021clip}, a large vision-language model, aligns visual and textual embeddings, facilitating zero-shot recognition of arbitrary object categories.
SAM~\cite{kirillov2023sam} achieves impressive zero-shot segmentation, providing high-quality instance masks for any object prompt.
However, as these foundation models are predominantly image-based, it is essential to develop methods that enable transfer of the rich knowledge from 2D to the 3D domain. 

\PAR{Radiance Fields for Scene Understanding.}
Neural radiance fields (NeRFs) \cite{mildenhall2020nerf} have become a popular representation for novel view synthesis and 3D scene understanding.
Methods such as LERF \cite{kerr2023lerf} and OpenNeRF \cite{engelmann2024opennerf} have introduced language embeddings, for example, CLIP features, to enable text-driven queries by distilling semantic information from the 2D domain into the continuous 3D volume provided by NeRF.

More recently, explicit representations such as 3D Gaussian Splatting (3DGS) \cite{kerbl2023splatting} have gained prominence.
3DGS represents a scene as a set of Gaussian primitives, achieving NeRF-like view synthesis quality while preserving an interpretable point-based structure.
Building upon this explicit representation, methods like LangSplat \cite{qin2024langsplat} have further extended semantic querying capabilities to 3DGS by integrating language embeddings into Gaussian representations.
Approaches such as Gaussian Grouping \cite{ye2024gaussiangrouping}, EgoLifter \cite{gu2024egolifter}, and ClickGaussian \cite{choi2024click} utilize SAM-generated 2D masks to segment and group Gaussians into distinct object instances, demonstrating the adaptability of 3D Gaussians for instance-level scene understanding.

In this work, we specifically leverage the explicit nature and rendering capability of 3DGS to ensure consistent multi-view feature transfer, which is crucial for effectively aligning rich 2D semantic observations with explicit 3D representations.

\PAR{Open-Vocabulary 3D Instance Segmentation.}
Traditional 3D instance segmentation approaches, typically voxel or point cloud-based, often rely heavily on manually annotated datasets, limiting their ability to generalize beyond predefined categories \cite{schult2023mask3d,SunSPFormer2023,kolodiazhnyi2024oneformer3d,Kolodiazhnyi_2024_WACV_TopDownVoxel,Han_2020_CVPR_Voxel}. 

To address these limitations, recent methods leverage vision-language models like CLIP \cite{radford2021clip} to enable open-vocabulary 3D scene understanding \cite{takmaz2023openmask3d, peng2023openscene, nguyen2024open3dis, huang2025segment3d}.
Methods like OpenScene \cite{peng2023openscene} focus on semantic understanding but lack instance-level granularity.
OpenMask3D \cite{takmaz2023openmask3d} addresses this by utilizing pretrained Mask3D~\cite{schult2023mask3d} and SAM~\cite{kirillov2023sam} models for instance labeling, and aggregates CLIP-based language information for each 3D instance.
However, Mask3D was trained on ScanNet200 \cite{dai2017scannet} and therefore is not a class-agnostic model.
Segment3D \cite{huang2025segment3d} improves upon this by leveraging refined 2D SAM masks to extract confident masks for partial point clouds and performing self-supervised fine-tuning on full scenes, enhancing generalization to novel objects and categories.

\begin{figure*}
    \centering
    \includegraphics[width=\linewidth]{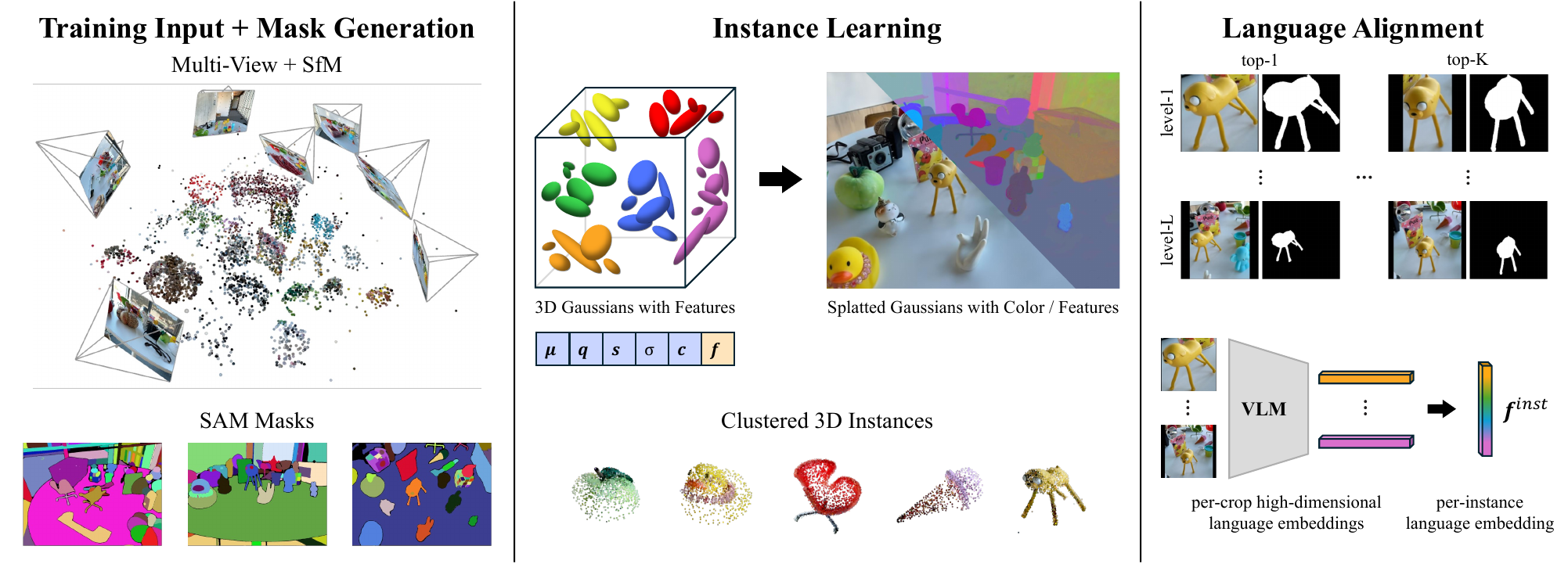}

    \caption{Overview of our proposed pipeline. On the left are the training inputs: posed RGB-images, a coarse SfM point cloud for initialization, and the extracted SAM masks. The middle section illustrates the instance learning with a Gaussian feature field optimization, as well as clustering to obtain coherent 3D instances. On the right, we demonstrate the language integration, where the top-k informative views are identified per instance, hierarchical crops are constructed and finally the language embedding per instance is computed.}
    \label{fig:method_overview}
\end{figure*}

Another line of work leveraging 3D Gaussian primitives as scene representation and exploiting the inherent 2D-to-3D connection is introduced by differentiable rendering.
Besides semantic \cite{qin2024langsplat} and instance segmentation methods \cite{ye2024gaussiangrouping, gu2024egolifter, silva2024contrastive}, the work most closely related to ours is OpenGaussian \cite{wu2024opengaussian}, which integrates instance-level features on 3D Gaussians with per-instance language embeddings to perform open-vocabulary 3D instance segmentation.
Both OpenGaussian and our method employ contrastive learning with similar positive-pair objectives, pulling features from the same instance together. However, while OpenGaussian penalizes negative pairs using inverse-distance weighting, our approach applies a margin-based loss to enforce a fixed minimum separation between distinct instances. Additionally, OpenGaussian explicitly enforces semantic consistency through a hierarchical discretized codebook, whereas our method implicitly encourages feature consistency via a novel variance-based regularization in continuous 3D space.
Furthermore, our method converges in fewer than half the iterations required by OpenGaussian, resulting in substantially accelerated optimization.

\section{Method}
\label{sec:method}
In this section, we introduce our approach for open-vocabulary 3D instance segmentation using 3D Gaussian representations.
An overview of our pipeline is shown in Figure \ref{fig:method_overview}.
In short, given a set of posed RGB images, we utilize the 3D Gaussian Splatting approach to optimize the 3D geometry of a scene and extend it to also optimize an instance embedding based on SAM masks.
This allows us to obtain a class-agnostic segmentation for the 3D scene.
These segments are in turn assigned a language feature based on a VLM output, which allows us to create a 3D open-vocabulary segmentation.
First, we provide essential background on 3D Gaussian Splatting to contextualize our work.
We then detail our method, and how we leverage the explicit and differentiable nature of 3DGS to effectively integrate semantic information from 2D observations into a coherent, multi-view consistent 3D representation.

\subsection{Background: 3D Gaussian Splatting}

3D Gaussian Splatting (3DGS) represents a scene explicitly using oriented 3D Gaussian primitives. Each Gaussian point $n$ corresponds to the center of a local 3D Gaussian distribution and is parameterized by $\theta_n = (\vect{\mu}_n, \mat{R}_n, \vect{s}_n, \vect{c}_n, o_n)$, where $\vect{\mu}_n$ is the 3D mean and $\mat{\Sigma}_n = (\mat{R}_n\mat{S}_n)^T\mat{S}_n\mat{R}_n$ is the covariance matrix parameterized via scale $\mat{S}_n=\text{diag}(\vect{s}_n)$ and orientation $\mat{R}_n$.
The view-dependent color is modeled via spherical harmonics $\vect{c}_n = \text{SH}_i(\vect{d}_n)$, with $\vect{d}_n$ denoting the view direction.
Additionally, each Gaussian has an associated scalar opacity $o_n$.

Rendering these Gaussians involves projecting them onto the 2D viewing plane via an approximation of the projective transformation,
generating 2D Gaussians termed \emph{splats}~\cite{kerbl2023splatting}.
The 2D covariance matrix of each projected Gaussian is given by:
\begin{equation}
    \mat{\hat\Sigma}_n = [\mat{J}_n \mat{R}_c \mat{\Sigma}_n \mat{R}_c^T \mat{J}_n^T]_{2\times 2}
\end{equation}
where $\mat{J}_n$ is the Jacobian of the transformation, and $\mat{R}_c$ is the camera's rotation matrix.
These splats are depth-sorted relative to the camera viewpoint and composited via alpha blending.
The contribution of a Gaussian at a pixel position $\vect{p}$ is determined by:
\begin{equation}
    \alpha_n = o_n \exp\left(-\frac{1}{2}(\vect{p} - \vect{\hat\mu}_n)^\top \mat{\hat\Sigma}_n^{-1} (\vect{p} - \vect{\hat\mu}_n)\right),
\end{equation}
where $\vect{\hat\mu}_n$ is the projected 2D mean.
The final pixel color intensity $\mat{I}(\vect{p})$ at pixel $\vect{p}$ is computed as:
\begin{equation}
    \mat{I}(\vect{p}) = \sum_{n=1}^N \vect{c}_n \alpha_n \prod_{j=1}^{n-1} (1 - \alpha_j),
\end{equation}
where $\vect{c}_n$ denotes the color contribution of Gaussian $n$, $\alpha_n$ the Gaussian's contribution and the product reflects the cumulative transmittance.

All rendering operations are implemented in an efficient CUDA-based rasterizer, allowing for rapid optimization and real-time rendering of complex scenes.
The rendering is fully differentiable, allowing optimization using standard gradient-based methods.
The Gaussians are optimized by minimizing image reconstruction errors with respect to a set of posed RGB images.
Specifically, the RGB rendering loss is defined as:
\begin{equation}
    \mathcal{L}_{\text{RGB}} = (1 - \beta) \mathcal{L}_{l_1} + \beta \mathcal{L}_{\text{SSIM}},
\end{equation}
where $\mathcal{L}_{l_1}$ denotes the $l_1$ loss for pixel-wise accuracy, $\mathcal{L}_{\text{SSIM}}$ the structural similarity loss for perceptual quality~\cite{ssim2004}, and $\beta$ is a weighting factor balancing the two terms.

Once optimized, scenes can be rendered from arbitrary viewpoints, facilitating real-time novel view synthesis.
Compared to conventional point clouds, Gaussian primitives can be  densely rendered without occlusion artifacts, and the differentiable rendering pipeline inherently links 2D supervision signals to the 3D representation, enabling effective 2D-to-3D correspondence.
These properties make 3D Gaussians Splatting highly suitable for integrating rich 2D information into detailed 3D representations.

\subsection{OpenSplat3D}

Given a set of posed RGB images $\{\mat{I}_v | v = 1,\ldots,V\}$ of a scene from multiple viewpoints $v$, we use the Segment Anything Model \cite{kirillov2023sam} to extract instance masks $\{ M_{v,i} | i = 1,\ldots,N_v\}$ for each image where $N_v$ represents the number of instances in viewpoint $v$.
We write $\vect{p} \in M_{v,i}$ to indicate that pixel $\vect{p}$ is part of instance $i$ in view $v$, and $|M_{v,i}|$ to denote the number of pixels belonging to instance $i$ in view $v$.
Unlike in Gaussian Grouping~\cite{ye2024gaussiangrouping}, these instance masks are not required to be consistent across different viewpoints.

To encode instance-level information within the 3D Gaussians, we extend each Gaussian primitive with a view-independent instance feature embedding $\vect{f}_n \in \mathbb{R}^d$.
These feature embeddings can be rendered from arbitrary viewpoints using the 3DGS splatting algorithm.
To be precise, the rendered feature map $\mat{F}(\vect{p})$ at pixel $\vect{p}$ is computed as:
\begin{equation}
    \mat{F}(\vect{p}) = \sum_{n=1}^N \bm{f}_n \alpha_n \prod_{j=1}^{n-1} (1 - \alpha_j),
\end{equation}
where $\mat{F} \in \mathbb{R}^{H \times W \times d}$.
This encoding provides two key benefits: First, the embeddings can be supervised with 2D data in 3D space via differentiable rendering.
Second, the structured feature embeddings in 3D space facilitate scene understanding tasks such as 3D instance segmentation.

\subsubsection{Instance Learning}
The per-Gaussian instance feature embeddings $\vect{f}_n \in \mathbb{R}^d$ are optimized by rendering a feature map from a given viewpoint $v$ and guiding the features using the 2D instance masks $M_{v,i}$ with a contrastive loss formulation, similar to prior work \cite{gu2024egolifter, silva2024contrastive, choi2024click}.
Given a rendered feature map, we compute a prototype feature $\vect{z}_i$ per instance mask:
\begin{equation}
    \vect{z}_i = \frac{1}{|M_{v,i}|} \sum_{\vect{p} \in M_{v,i}} \mat{F}(\vect{p}).
\end{equation}
This is used in the positive part of the contrastive loss to pull all features of the corresponding mask to the prototype by:
\begin{equation}
    \mathcal{L}_{\text{pos}} = \frac{1}{N_v} \sum_{i=1}^{N_v} \frac{1}{|M_{v,i}|} \sum_{\vect{p} \in M_{v,i}} \lVert \mat{F}(\vect{p}) - \vect{z}_i \rVert_2^2,
\end{equation}
where $|M_{v,i}|$ denotes the number of pixels in the mask for instance $i$ and viewpoint $v$.
The negative part of the contrastive loss pushes the prototypes away from each other using a margin-based loss:
\begin{equation}
    \mathcal{L}_{\text{neg}} = \frac{2}{N_v(N_v - 1)} \sum_{i=1}^{N_v} \sum_{j > i}^{N_v} \text{ReLU}(\gamma - \lVert \vect{z}_i - \vect{z}_j \rVert_2^2),
\end{equation}
where $\gamma$ is the margin, a hyperparameter which defines the boundary where negative prototypes are too close and should be pushed away, preventing embeddings from collapsing into a single cluster. 

The final contrastive instance-level loss is defined as:
\begin{equation}
    \mathcal{L}_{\text{inst2D}} = w_p \cdot \mathcal{L}_{\text{pos}} + w_n \cdot \mathcal{L}_{\text{neg}},
\end{equation}
where $w_p$ and $w_n$ control the balance between instance compactness and separability, ultimately enhancing clusterability.
While the SAM segmentation masks for the different views might give somewhat conflicting optimization targets, we empirically find that SAM instance masks across a scene typically tend to have a sufficiently consistent mode of segmenting objects. The joint optimization across the different views can be seen as a sort of majority voting during optimization and merely serves as a supervisory signal for the instance embeddings.

After optimization, the instance embeddings $\vect{f}_n$, and hence the Gaussians, are clustered into meaningful groups while also determining the number of clusters and identifying outliers.
We leverage HDBSCAN \cite{hdbscan2013}, which is noise-aware and probabilistic, making it well-suited for the inherent noise and uncertainty in the instance labels and Gaussian representation.

\subsubsection{Language Alignment}
Previous works without explicit instance information \cite{qin2024langsplat, kerr2023lerf} compute per-point language embeddings.
Specifically within the context of 3DGS, it is infeasible to optimize high dimensional instance information for each Gaussian.
For this reason, LangSplat projects language features to a low-dimension space with a scene specific auto-encoder~\cite{qin2024langsplat}.
To avoid this, we follow recent approaches and adopt a per-instance embedding strategy inspired by \cite{takmaz2023openmask3d, huang2025segment3d, wu2024opengaussian}.

Rather than assigning and optimizing a language embedding for each Gaussian, we generate a single language embedding per instance after the optimization.
To obtain this instance-level embedding, we render binary masks for each instance from multiple viewpoints.
Gaussians belonging to a specific instance are rendered in white, while all others are set to black, resulting in an object silhouette respecting occlusion.
Afterwards, the rendering is thresholded to produce a binary mask.
For selecting the most informative views, we employ a visibility-based scoring method similar to OpenMask3D~\cite{takmaz2023openmask3d}.
The visibility score for a given instance $i$ and a viewpoint $v$ is computed as:
\begin{equation}
    s_{v,i} = \frac{|\mat{M}_{v,i}|}{|\mat{I}_v|} \cdot \frac{|G_{v,i}|}{|G_i|},
\end{equation}
where $|\mat{I}_v|$ is the number of pixels of image $\mat{I}_v$, $|G_i|$ is the number of Gaussians associated with instance $i$, and $|G_{v,i}|$ represents the number of Gaussians for instance $i$  visible in the viewport of viewpoint $v$.
Since the rasterizer provides information about all Gaussians within the viewport, $|G_{v,i}|$ can be computed efficiently.

For each of the selected top-$K$ viewpoints, $L$ image crops centered around the predicted instance mask are extracted at different zoom levels to capture both detailed and contextual information.
These crops are processed by the image encoder of a vision-language model to produce language embeddings $\vect{l}_{i,k,l}$, where $i$ indicates the instance, $k$ represents the viewpoint, and $l$ denotes the zoom level.

The final language embedding for the instance is obtained by averaging the embeddings across the selected viewpoints and zoom levels:
\begin{equation}
    \bm{l}_i = \frac{1}{KL} \sum_{k=1}^K \sum_{l=1}^L \vect{l}_{i,k,l}.
\end{equation}
This approach ensures that the language embedding captures diverse visual information about the instance, enhancing its semantic representation.

\subsubsection{Instance Feature Regularization}

A significant challenge in the differential rendering via alpha compositing lies in the discrepancy between 2D supervision and the underlying 3D representation.
During rendering, overlapping Gaussians are blended, combining their features to produce the final 2D feature map.
Although this blending ensures alignment with 2D supervision signals, it does not inherently enforce consistency among the individual Gaussian embeddings. Consequently, they can diverge, potentially leading to inconsistent instance representations or feature misalignment across views.

To address this, we propose a novel variance regularization loss that minimizes the variance of feature embeddings along rendering rays.
Using $\Var(X) = E(X^2) - E(X)^2$, we compute the variance of the feature map $\mat{F}$ at pixel $\vect{p}$ as
\begin{equation}
    \Var(\mat{F}(\vect{p})) = \left( \sum_{n=1}^N \bm{f}_n^2 \alpha_n \prod_{j=1}^{n-1} (1 - \alpha_j) \right)- \mat{F}(\vect{p})^2,
\end{equation}
which is efficiently calculated within the CUDA kernel during rendering, minimizing computational overhead.
The variance is minimized by an $l_2$ loss given as
\begin{equation}
    \mathcal{L}_{\text{var}} = \frac{1}{|\mat{I}|} \sum_{\vect{p} \in \mat{I}} \lVert \Var(\mat{F}(\vect{p})) \rVert_2^2.
\end{equation}
Our final optimization objective combines the RGB reconstruction loss with the contrastive instance loss and the variance regularization term:
\begin{equation}
    \mathcal{L} = \mathcal{L}_{\text{RGB}} + \lambda_{\text{inst2d}} \mathcal{L}_{\text{inst2d}} +  \lambda_{\text{var}} \mathcal{L}_{\text{var}}, \end{equation}
where $\lambda_{\text{inst2d}}$ and $\lambda_{\text{var}}$ are weighting factors of each loss.
\section{Experiments}
\label{sec:experiments}

\begin{table*}[ht!]
    \centering
\begin{tabularx}{\linewidth}{lcYYcYYcYYcYY}
\toprule
\multirow{2}[2]{*}{Method} && \multicolumn{2}{c}{figurines} && \multicolumn{2}{c}{ramen}                     && \multicolumn{2}{c}{teatime}           && \multicolumn{2}{c}{\textbf{mean}} \\
\cmidrule{3-4}
\cmidrule{6-7}
\cmidrule{9-10}
\cmidrule{12-13}
                        && mIoU          & mBIoU         && mIoU                  & mBIoU                 && mIoU          & mBIoU                 && mIoU                  & mBIoU\\
\midrule
LERF~\cite{kerr2023lerf}*                   && 33.5        & 30.6        && 28.3                & 14.7                && 49.7        & 42.6                && 37.2                & 29.3 \\
LangSplat~\cite{qin2024langsplat}*              && 52.8        & 50.5        && 50.4                & 44.7                && 69.5        & 65.6                && 57.6                & 53.6 \\
\midrule
Gaussian Grouping~\cite{ye2024gaussiangrouping}*      && 69.7        & 67.9        && \textbf{77.0}           & \textbf{68.7}           && 71.7        & 66.1                && 72.8                & 67.6 \\
CGC~\cite{silva2024contrastive}                     && \underline{91.6}        & \underline{88.8}        && 68.7                & 63.1                && \underline{80.5}        & \textbf{78.9}           && \underline{80.3}    & \underline{76.9} \\
\midrule
OpenSplat3D (Ours)           && \textbf{92.3}   & \textbf{89.4}   && \underline{75.9}    & \underline{68.2}    && \textbf{83.7}   & \underline{78.8}    && \textbf{84.0}           & \textbf{78.8} \\ \bottomrule
\end{tabularx}
\vspace{-5pt}
    \caption{Semantic segmentation results on the LERF-mask dataset. We report the mean IoU and mean BIoU for each scene and the overall average. Our method achieves the best overall performance across all metrics. Only for the ramen scene, Gaussian Grouping performs slightly better. *: Results as reported in the Gaussian Grouping paper \cite{ye2024gaussiangrouping}.}
    \label{tab:lerf_mask_eval}
\end{table*}

\begin{table*}[ht!]
    \centering

\begin{tabularx}{\linewidth}{p{2.9cm}YYcYYcYYcYYcYY}
\toprule
\multirow{2}[2]{*}{Method} & \multicolumn{2}{c}{figurines} && \multicolumn{2}{c}{ramen} && \multicolumn{2}{c}{teatime} && \multicolumn{2}{c}{waldo\_kitchen} &~~~~& \multicolumn{2}{c}{\textbf{mean}} \\
\cmidrule{2-3}
\cmidrule{5-6}
\cmidrule{8-9}
\cmidrule{11-12}
\cmidrule{14-15}
& mIoU & mAcc. && mIoU & mAcc. && mIoU & mAcc. && mIoU & mAcc. && mIoU & mAcc. \\
\midrule
LangSplat~\cite{qin2024langsplat} & 10.16 & ~~8.93 && ~~7.92 & 11.27 && 11.38 & 20.34 && ~~9.18 & ~~9.09 && ~~9.66 & 12.41 \\
LEGaussians~\cite{shi2024legaussian} & 17.99 & 23.21 && 15.79 & 26.76 && 19.27 & 27.12 && 11.78 & 18.18 && 16.21 & 23.82 \\
OpenGaussian~\cite{wu2024opengaussian} & 39.29 & 55.36 && 31.01 & 42.25 && 60.44 & 76.27 && 22.70 & 31.82 && 38.36 & 51.43 \\
\midrule
OpenSplat3D (Ours) & \textbf{60.71} & \textbf{85.71} && \textbf{49.20} & \textbf{76.06} && \textbf{73.27} & \textbf{88.14} && \textbf{55.63} & \textbf{77.27} && \textbf{59.70} & \textbf{81.79}\\

\bottomrule
\end{tabularx}
    \vspace{-5pt}
    \caption{LERF-OVS 3D object selection evaluation from textual query. Following OpenGaussian~\cite{wu2024opengaussian}, only the Gaussians responding to the query are rendered, therefore the rendering does not respect occlusion by other objects in the scene. Accuracy is provided by mAcc@$0.25$. Note that OpenGaussian fine-tunes parameters per scene for best results.}
    \label{tab:lerf_ovs_eval}
\end{table*}

\PARbegin{Implementation Details.}
For the instance mask extraction, we utilize only the default output from SAM~\cite{kirillov2023sam}, sorting the masks by the combination of intersection-over-union (IoU) and stability scores and combining the binary masks into a single mask containing unique instance IDs.

We retain most hyperparameters of the original Gaussian Splatting formulation~\cite{kerbl2023splatting}.
Since our additional loss terms lead to higher per-Gaussian gradients, we increase the gradient threshold for densification, preventing excessive growth in the number of Gaussians during optimization.
Following 3DGS, each scene is optimized for 30000 iterations.
For the instance feature embedding $\vect{f}_n^i \in \mathbb{R}^d$ we set the dimension to $d = 8$ in all our experiments.
Loss weights are set to $\lambda_{\text{inst2d}} = 0.1$, $\lambda_{\text{var}} = 0.5$, $w_{\text{pos}} = 1.0$, $w_{\text{neg}} = 1.0$, and the margin of the negative contrastive loss to $\gamma=1.0$.
Notably, we employ a consistent set of hyperparameters across all experiments, in contrast to OpenGaussian \cite{wu2024opengaussian} which uses adapted parameters per scene to achieve optimal results. The experiments are conducted on an RTX 3090 GPU. Optimizing the instances per scene typically takes 20-45 minutes.
Clustering via the cuML~\cite{raschka2020cuml} CUDA implementation of HDBSCAN takes between a few seconds and approximately 2 minutes depending on the number of Gaussians.
For the language embeddings, we use MasQCLIP \cite{xu2023masqclip} as vision-language model due to its usage of masks to generate embeddings for specific objects in an image while retaining context. The computation of these embeddings using top $k=5$ views and $l=3$ zoom levels with an expansion ratio of $0.3$ finishes in a few minutes, varying based on the number of instances in the scene.

\PAR{Datasets.}
We evaluate our method on open-vocabulary segmentation on two datasets, LERF-mask~\cite{ye2024gaussiangrouping} and LERF-OVS.
Both datasets extend the original LERF dataset introduced by Kerr \etal~\cite{kerr2023lerf}, which provides multi-view images of real-world scenes with accurate camera poses and annotations for the task of object localization.
LERF-mask and LERF-OVS augment a subset of these scenes with additional semantic segmentation masks and corresponding textual prompts for various objects.

Furthermore, to comprehensively assess the ability of 3D instance segmentation of our method, we perform evaluations on the ScanNet++ (v1) dataset \cite{yeshwanth2023scannet++}, utilizing its validation split consisting of 50 annotated indoor scenes.
Note that this dataset actually performs the evaluation in 3D, unlike the LERF-Mask/OVS datasets which evaluate the performance on 2D views. Additionally, ScanNet++ is significantly larger, is annotated much more densely, and evaluated across 84 instance classes.

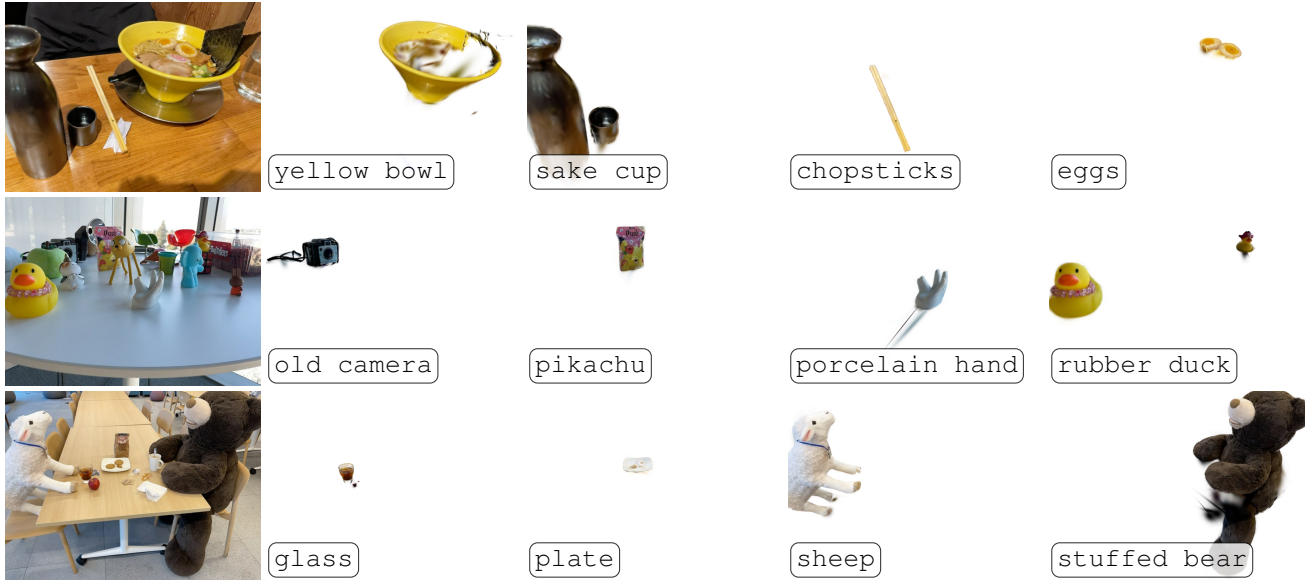
\begin{figure*}[ht]
    \centering

    \begin{tikzpicture}[
        image/.style={
inner sep=0pt,
            outer sep=1pt,
},
        label/.style={
            anchor=south west, 
            outer sep=2pt,
inner sep=2pt,
            color=black, 
            fill=white,
            opacity=0.8,
            text opacity=1,
            rounded corners=3pt,
            draw=black,
            text height=1.5ex,
            text depth=0.5ex,
        },
        pics/splitImage/.style 2 args={code={\node[image] (-img1) at (0, 0) {#1};\begin{scope}\clip ($(-img1.north west)!0.333!(-img1.north east)$)
    				-- (-img1.north east)
    				-- (-img1.south east)
    				-- ($(-img1.south west)!0.667!(-img1.south east)$)
    				-- cycle;\node[image] (-img2) at (0, 0) {#2};\end{scope}}},
    ]

    \node[image] (img) {\includegraphics[width=3.4cm]{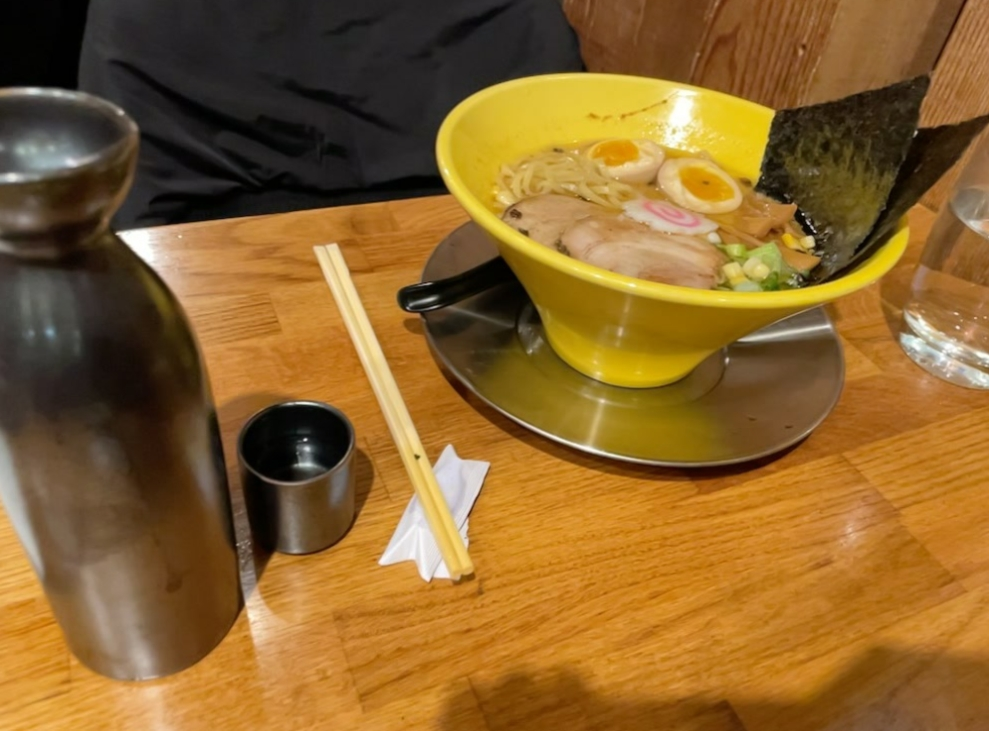}};

    \node[image, anchor=west] (bowl) at (img.east) {\includegraphics[width=3.4cm]{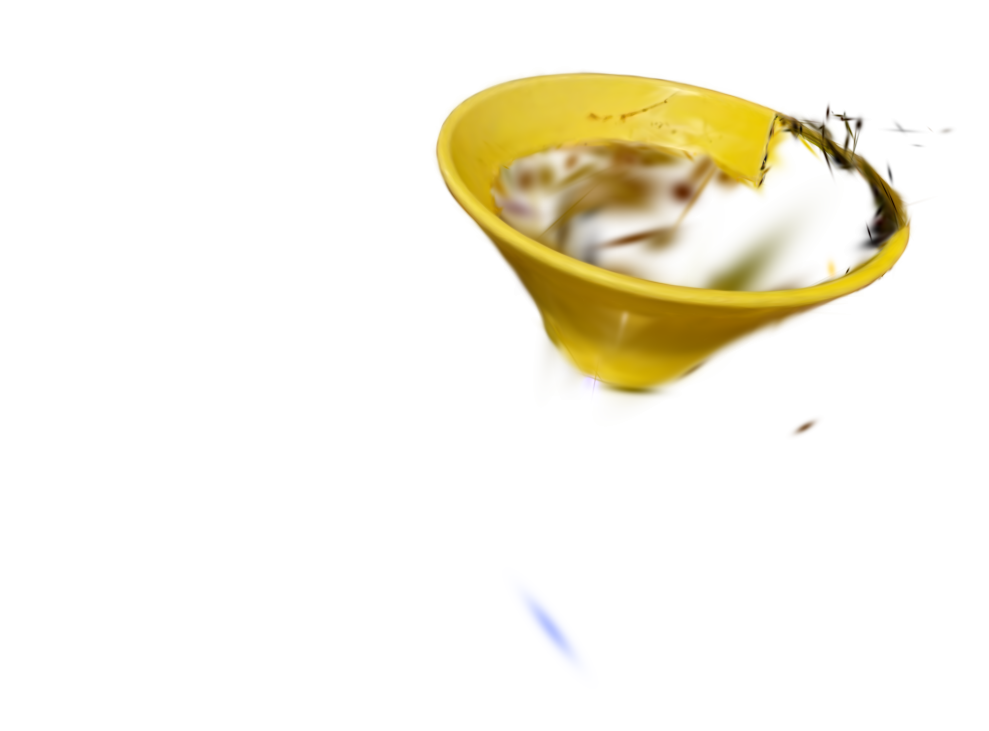}};
    \node[label] at (bowl.south west) {\texttt{yellow bowl}};

    \node[image, anchor=west] (sake) at (bowl.east) {\includegraphics[width=3.4cm]{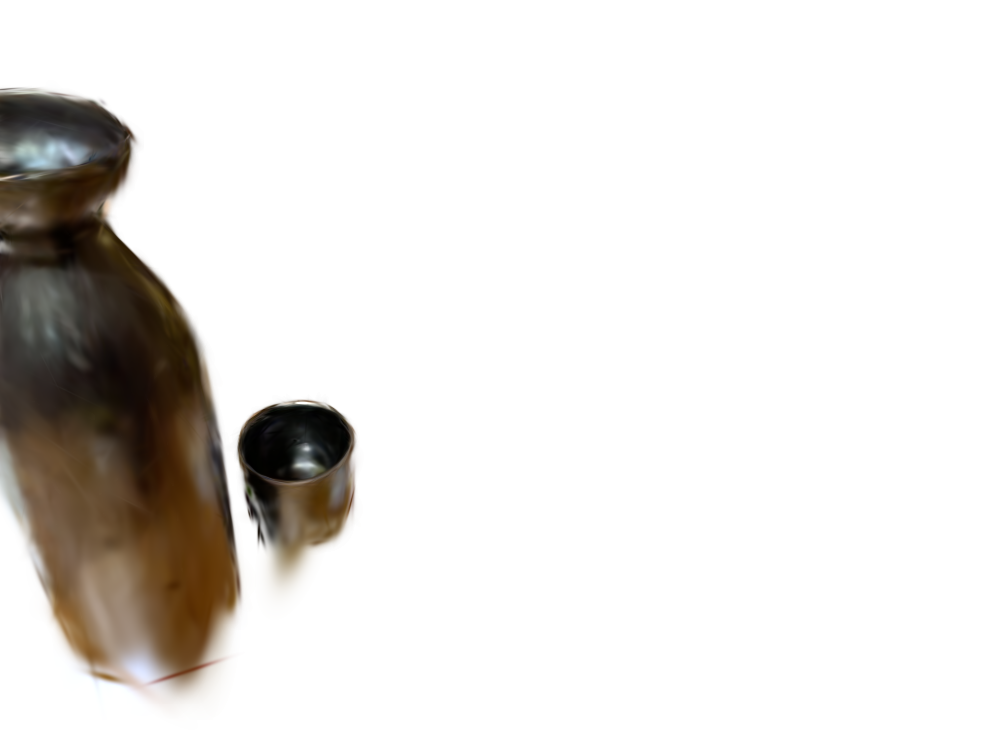}};
    \node[label] at (sake.south west) {\texttt{sake cup}};

    \node[image, anchor=west] (sticks) at (sake.east) {\includegraphics[width=3.4cm]{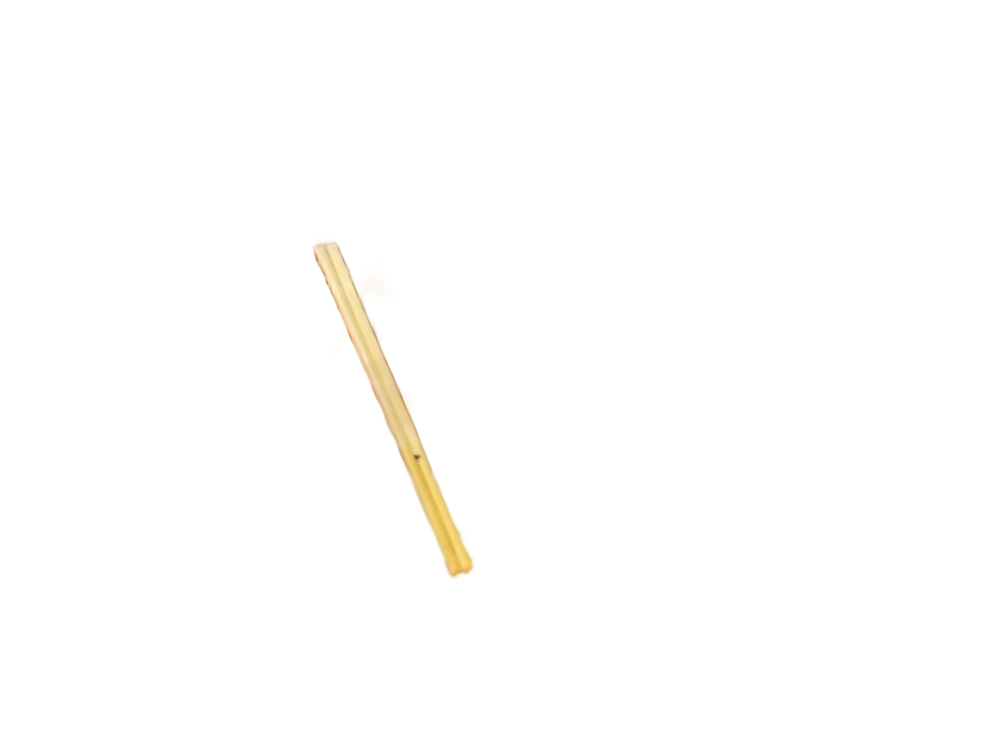}};
    \node[label] at (sticks.south west) {\texttt{chopsticks}};

    \node[image, anchor=west] (eggs) at (sticks.east) {\includegraphics[width=3.4cm]{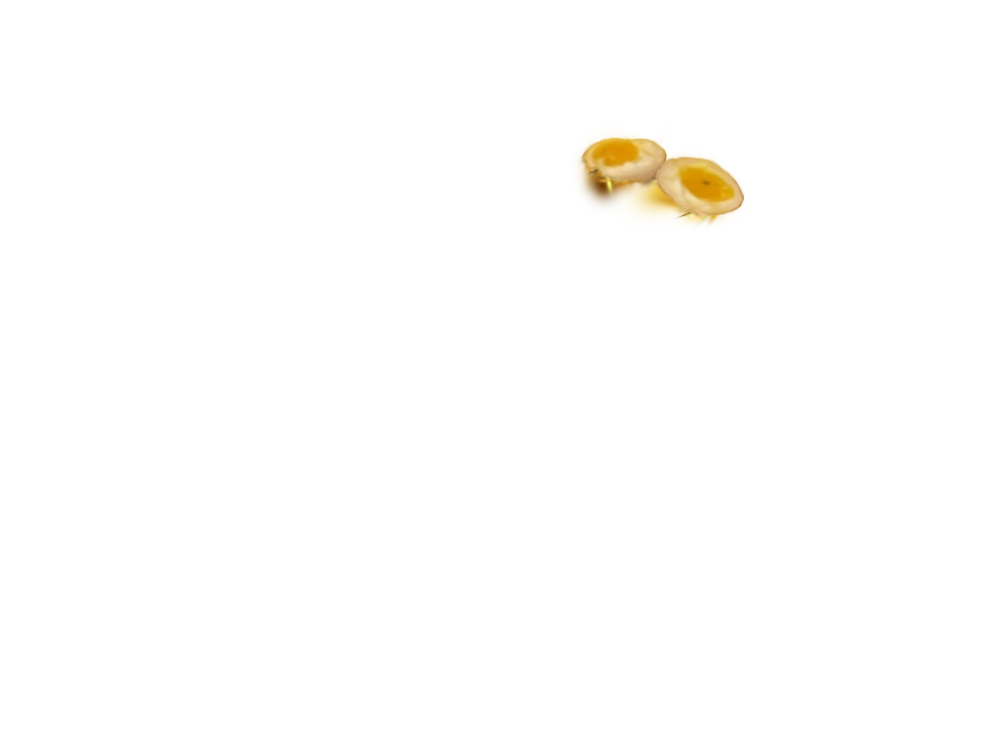}};
    \node[label] at (eggs.south west) {\texttt{eggs}};

    \node[image, anchor=north] (img2) at (img.south) {\includegraphics[width=3.4cm, height=3.5cm, keepaspectratio]{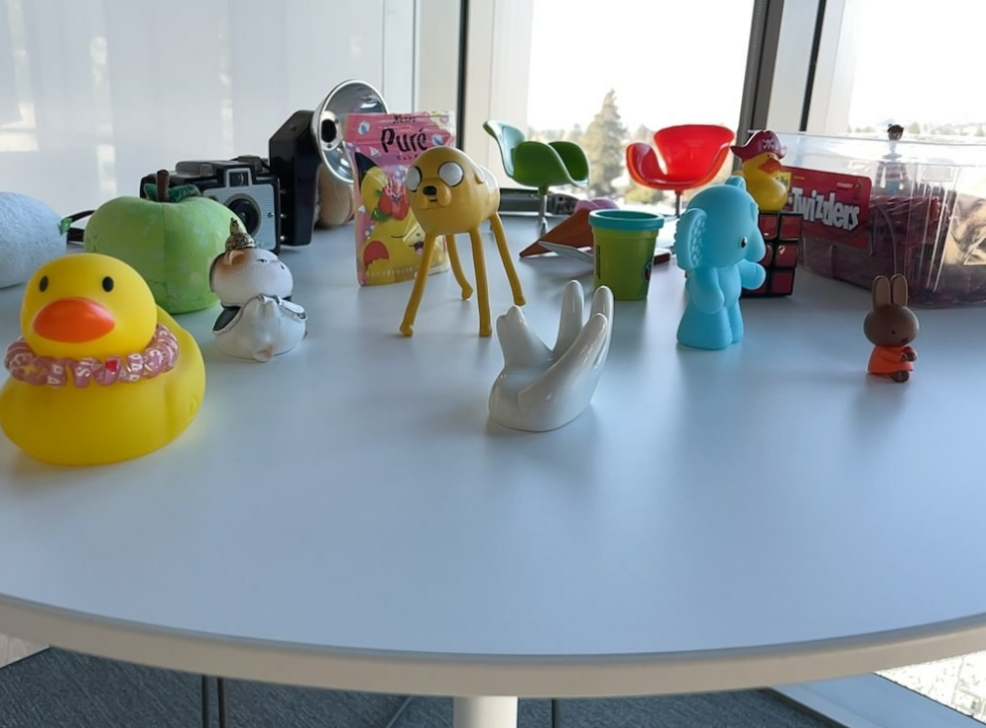}};

    \node[image, anchor=west] (bowl) at (img2.east) {\includegraphics[width=3.4cm, height=3.5cm, keepaspectratio]{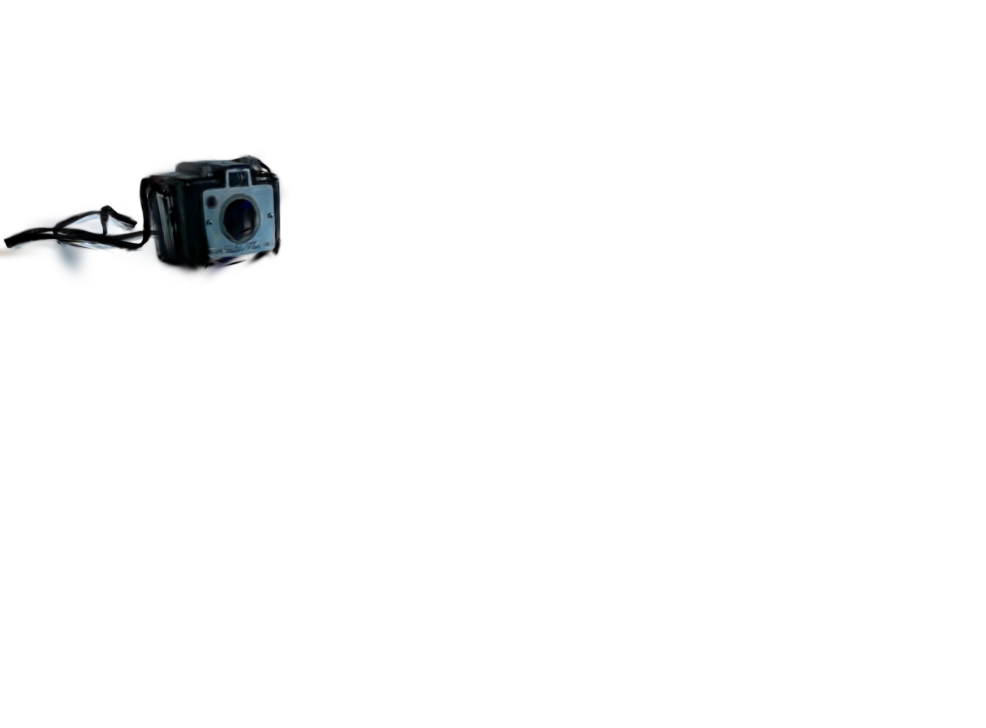}};
    \node[label] at (bowl.south west) {\texttt{old camera}};

    \node[image, anchor=west] (sake) at (bowl.east) {\includegraphics[width=3.4cm]{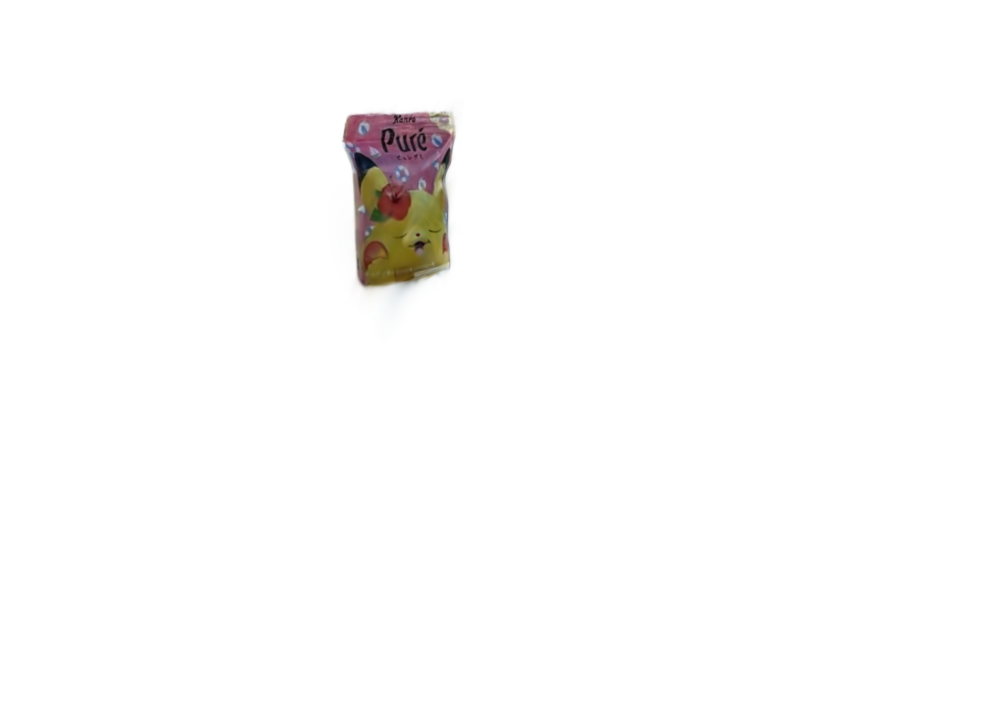}};
    \node[label] at (sake.south west) {\texttt{pikachu}};

    \node[image, anchor=west] (sticks) at (sake.east) {\includegraphics[width=3.4cm]{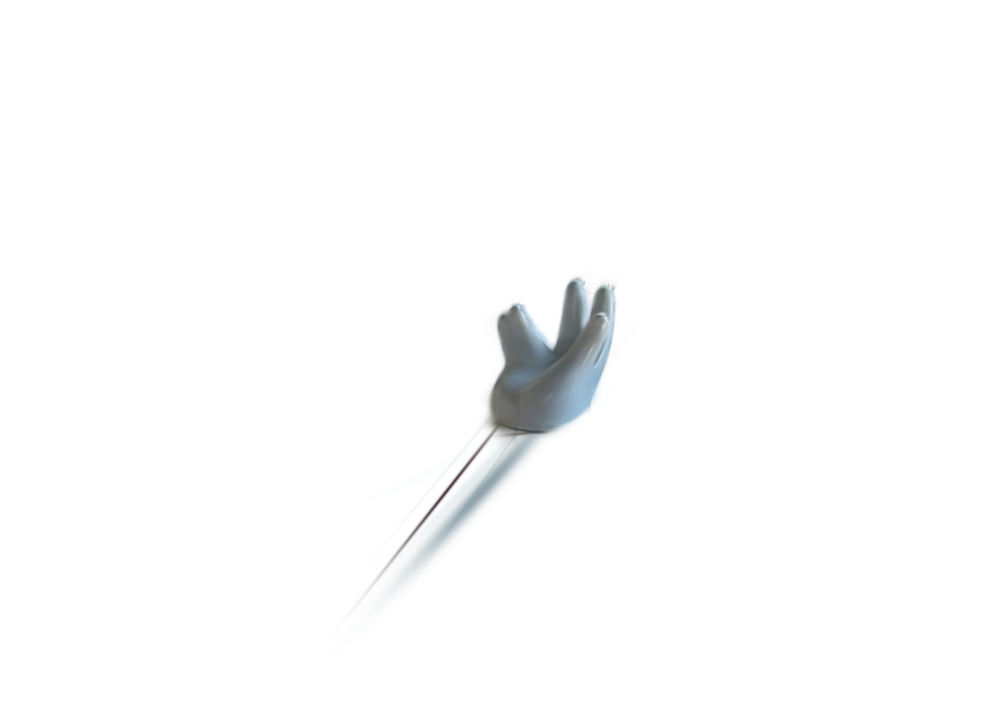}};
    \node[label] at (sticks.south west) {\texttt{porcelain hand}};

    \node[image, anchor=west] (eggs) at (sticks.east) {\includegraphics[width=3.4cm]{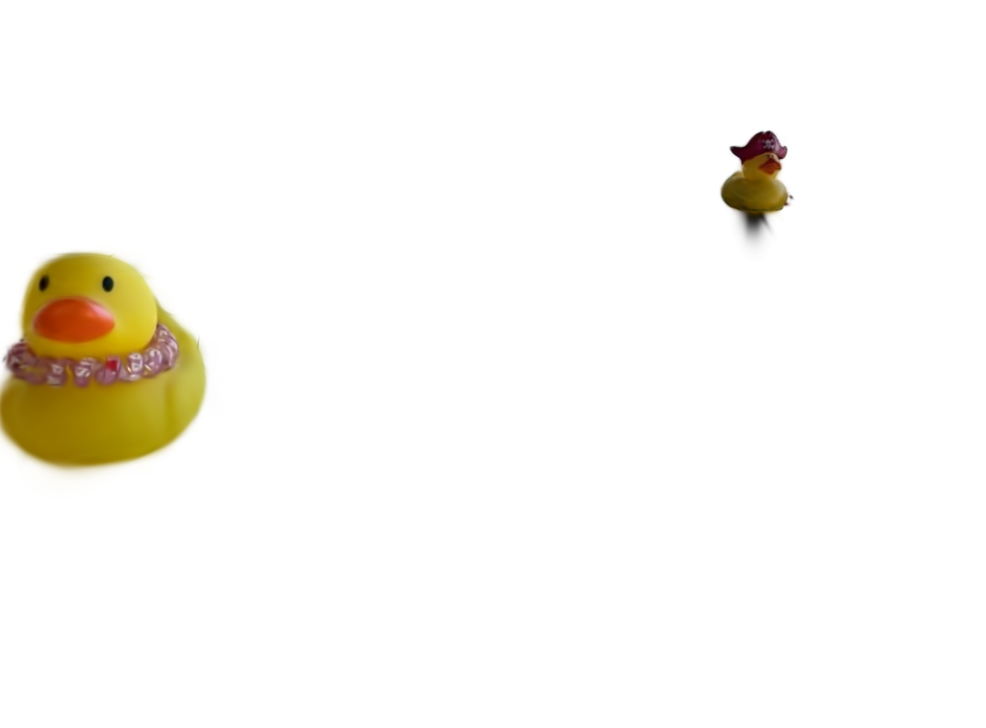}};
    \node[label] at (eggs.south west) {\texttt{rubber duck}};

    \node[image, anchor=north] (img3) at (img2.south) {\includegraphics[width=3.4cm, height=3.5cm, keepaspectratio]{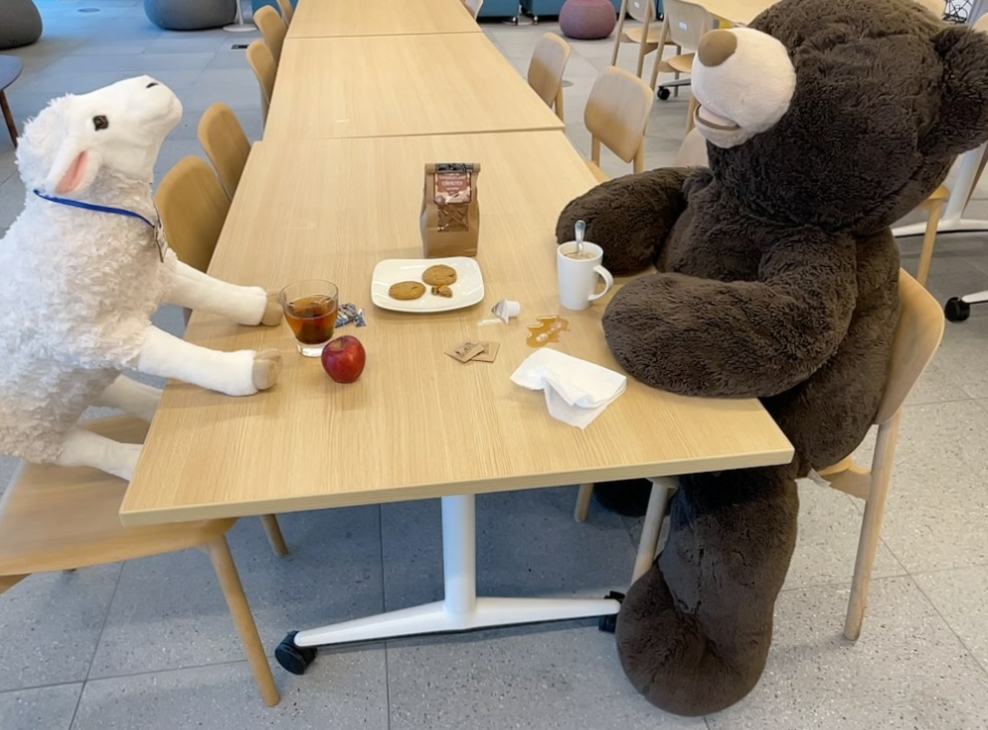}};

    \node[image, anchor=west] (bowl) at (img3.east) {\includegraphics[width=3.4cm, height=3.5cm, keepaspectratio]{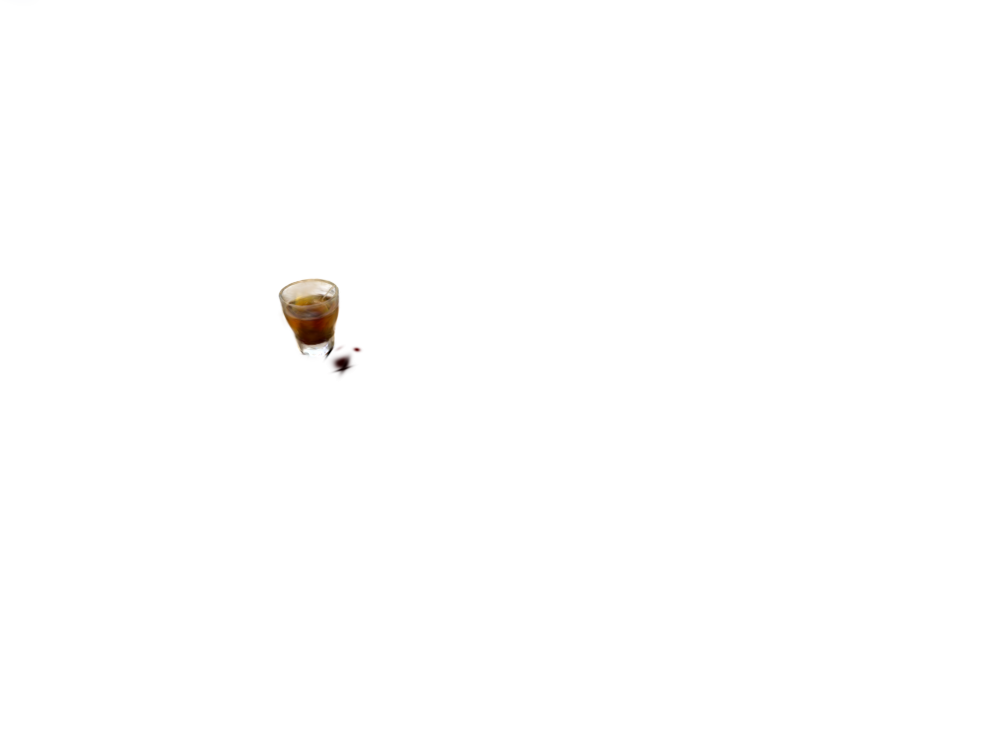}};
    \node[label] at (bowl.south west) {\texttt{glass}};

    \node[image, anchor=west] (sake) at (bowl.east) {\includegraphics[width=3.4cm]{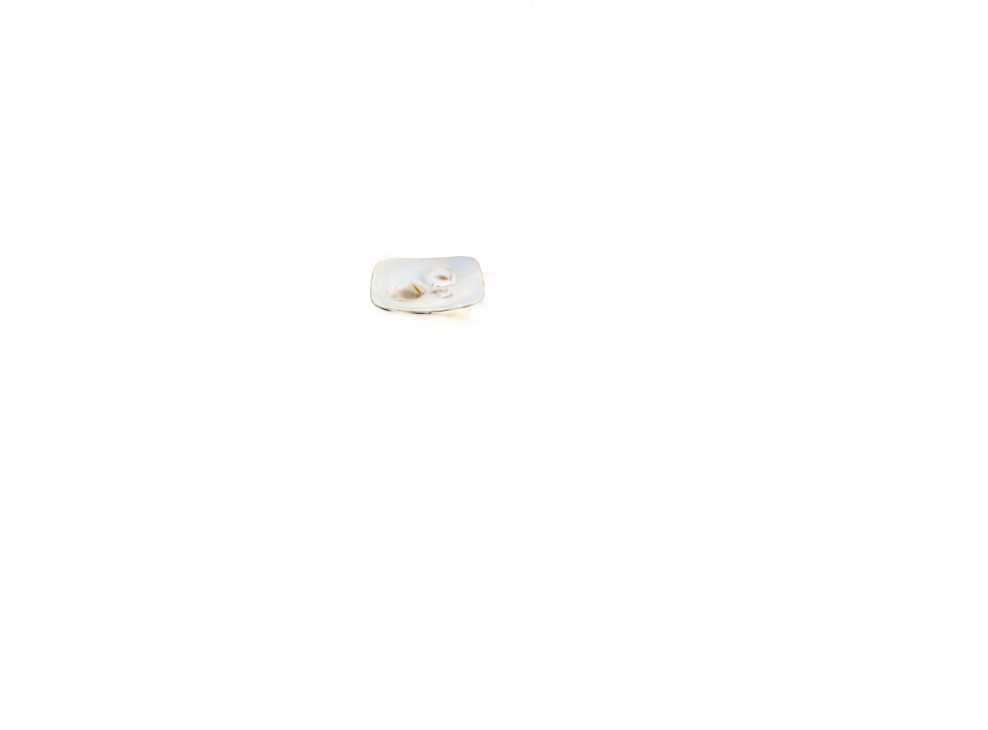}};
    \node[label] at (sake.south west) {\texttt{plate}};

    \node[image, anchor=west] (sticks) at (sake.east) {\includegraphics[width=3.4cm]{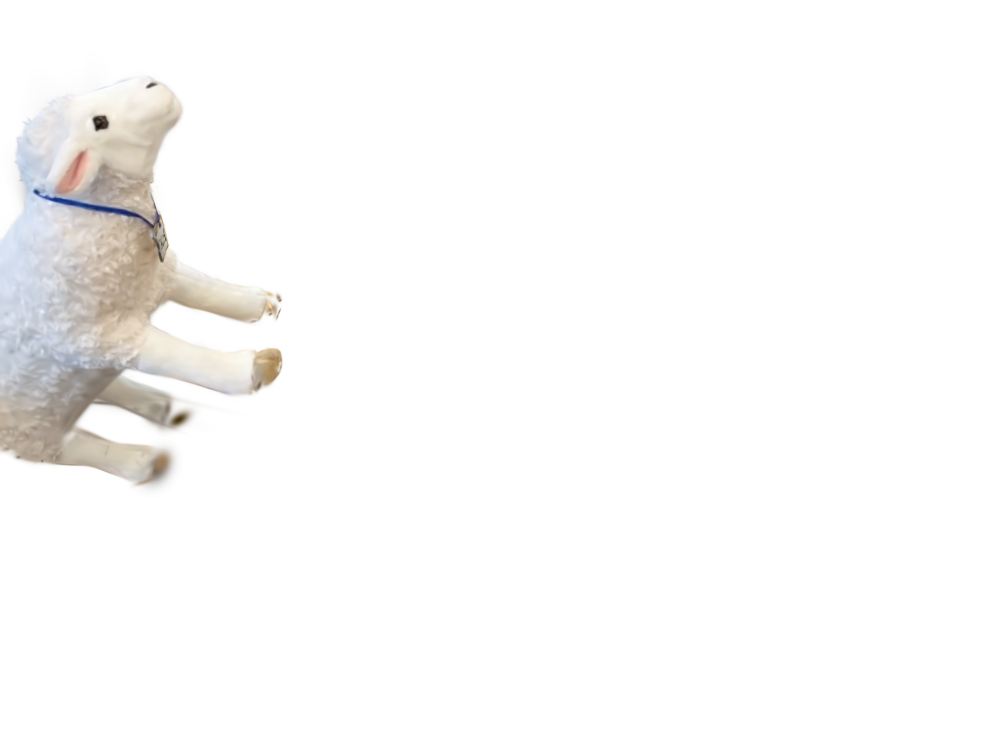}};
    \node[label] at (sticks.south west) {\texttt{sheep}};

    \node[image, anchor=west] (eggs) at (sticks.east) {\includegraphics[width=3.4cm]{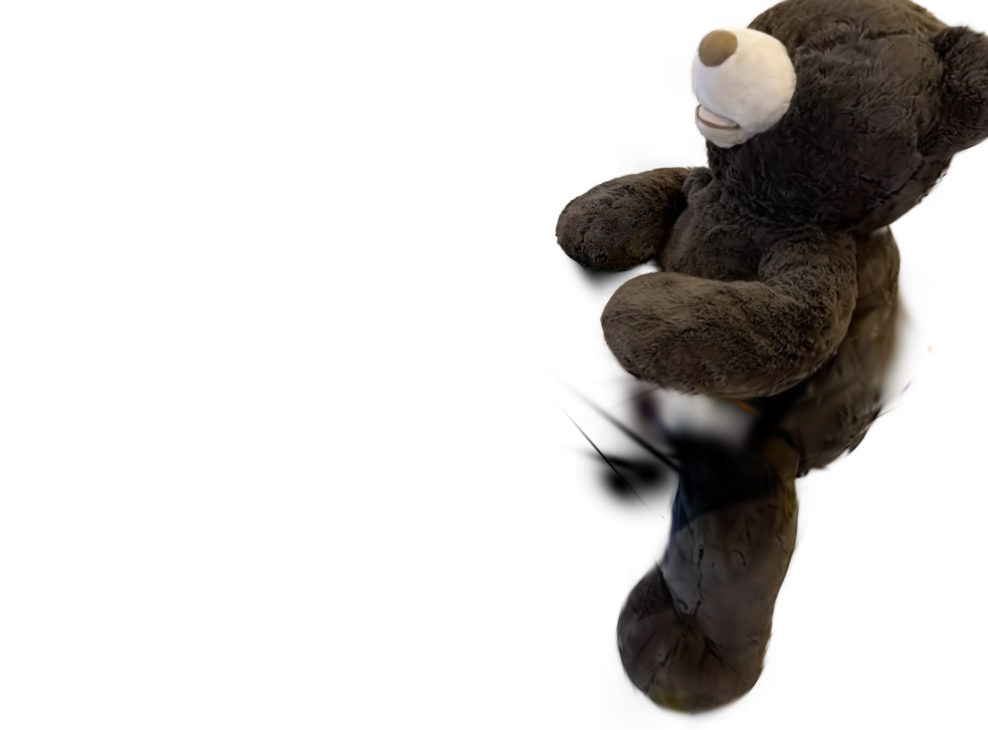}};
    \node[label] at (eggs.south west) {\texttt{stuffed bear}};
    
    \end{tikzpicture}
\caption{
    Language query results on the LERF-mask dataset.
    The left side shows example RGB training images for reference.
    The right side showcases rendered 3D instances extracted with OpenSplat3D using natural language queries.
    The query string is inset in each image.
    }\vspace{-5pt}
    \label{fig:qualitative_lerf_mask_language_object_extraction}
\end{figure*}

\subsection{Open-Vocabulary Segmentation}

\paragraph{LERF-mask.}

For a fair comparison with other approaches, we adopt the same observer-based protocol proposed by Gaussian Grouping~\cite{ye2024gaussiangrouping} for the LERF-mask dataset.
Given an observed reference frame and a textual prompt, the open-vocabulary detector Grounding DINO~\cite{liu2025grounding} is leveraged to compute object proposals in form of bounding boxes, which are subsequently used as prompts for SAM to extract semantic mask proposals.
These masks are used to determine the instances of interest by splatting a binary mask for each instance into the 2D view and selecting the instances over an intersection-over-area (\textit{IoA}) threshold.
In the subsequent evaluation frames, only the masks for the selected instances are rendered.
This multi-view evaluation strategy assesses the quality of 3D localization via intermediate 2D segmentation consistency, requiring the model to render stable and accurate 2D segmentation masks across multiple views based on the information of the reference frame.

Table \ref{tab:lerf_mask_eval} reports results on LERF-mask using the official metrics, specifically the mean Intersection-over-Union (\textit{mIoU}) for masks and bounding boxes (\textit{mBIoU}).
\ours shows superior performance on most scenes, achieving significantly better scores on average.
Gaussian Grouping performs slightly better on the \textit{ramen} scene.

We present qualitative results for textual queries and object extraction on the LERF-mask dataset in Figure~\ref{fig:qualitative_lerf_mask_language_object_extraction}. To clearly illustrate the Gaussians belonging to each instance, we render extracted objects without occlusion.
This visualization also highlights Gaussians incorrectly associated with the instances, typically found near object boundaries that merge with other objects, \eg, the porcelain hand with the table, or due to specular reflections.
These specularity-related Gaussians often have an offset from the surface due to the viewpoint-dependent nature of reflections.

\PAR{LERF-OVS.}
The LERF-OVS dataset, introduced by \cite{qin2024langsplat}, provides an open-vocabulary segmentation benchmark of the LERF dataset on four subscenes: \textit{figurines}, \textit{ramen}, \textit{teatime}, \textit{waldo\_kitchen}.
OpenGaussian \cite{wu2024opengaussian} introduced a protocol to evaluate 3D object selection on this dataset, differing from earlier observer-based evaluations that depended on 2D reference views. Instead, their protocol directly selects Gaussians based on language queries without preliminary rendering.
Methods such as LangSplat and LEGaussian utilize \emph{per-point} language embeddings which can be leveraged to select the Gaussians based on the similarity to the language.
In contrast, both OpenGaussian and our method employ \emph{per-instance} language embeddings, selecting entire instances based on embedding similarity.
Selected Gaussians are rendered to binary masks from multiple viewpoints without occlusion handling.
Although this rendering approach lowers performance metrics due to the occlusion-aware ground-truth masks, it effectively highlights Gaussians incorrectly assigned to instances in the 3D space. 
We evaluate our approach using the \textit{mIoU} and \textit{mAcc.} metrics for the 3D object selection task in Table \ref{tab:lerf_ovs_eval}.
Our method notably outperforms existing approaches---including the recent OpenGaussian---across all metrics, demonstrating significant improvements in 3D object selection accuracy.
We attribute this to our novel variance loss.

\subsection{3D Instance Segmentation}

\begin{figure*}[htbp]
  \centering

\begin{minipage}[b]{0.23\textwidth}
    \centering
    \includegraphics[width=\linewidth]{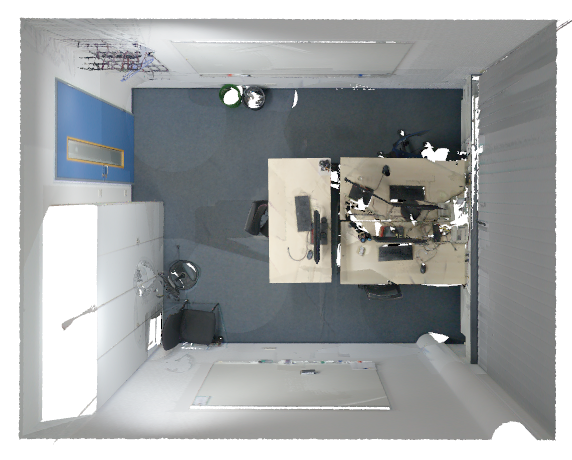} \\
    \includegraphics[width=\linewidth]{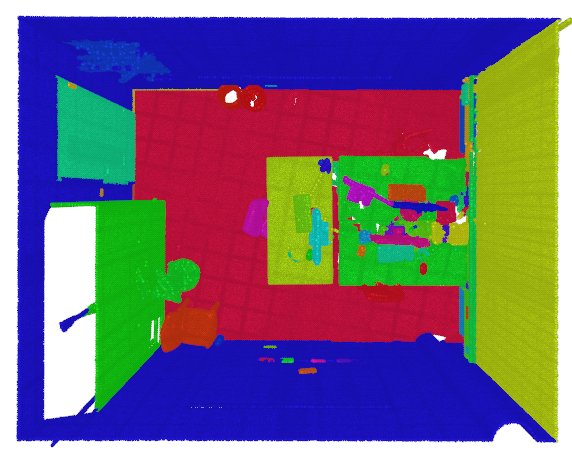} \\[-5pt]
    \caption*{1ada7a0617}
  \end{minipage}
  \hfill
\begin{minipage}[b]{0.18\textwidth}
    \centering
    \includegraphics[width=\linewidth]{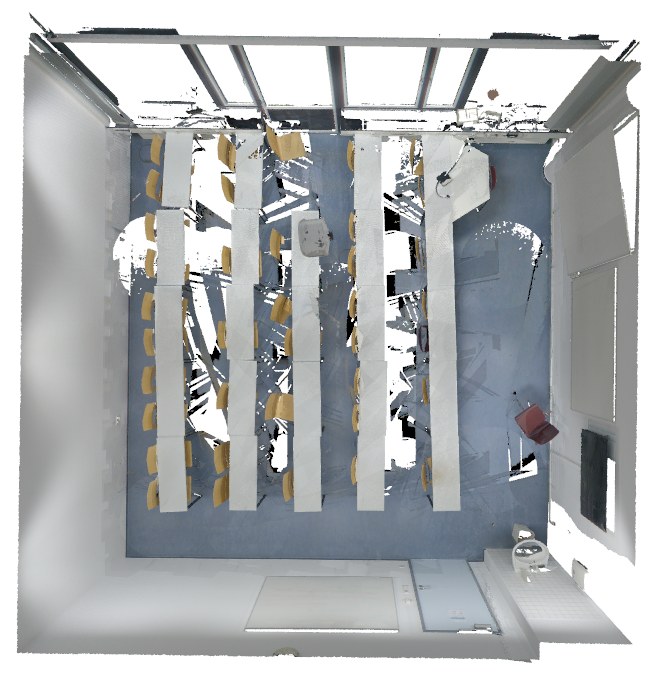} \\
    \includegraphics[width=\linewidth]{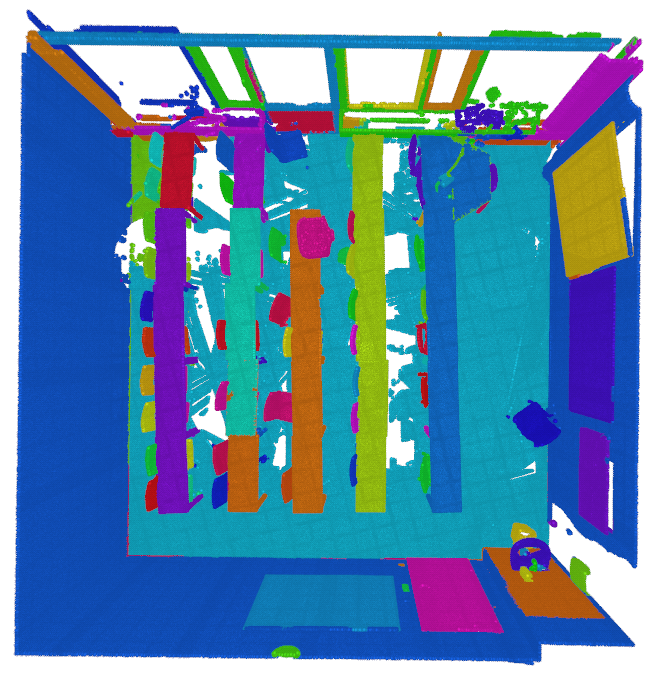} \\[-5pt]
    \caption*{21d970d8de}
  \end{minipage}
  \hfill
\begin{minipage}[b]{0.3\textwidth}
    \centering
    \includegraphics[width=\linewidth]{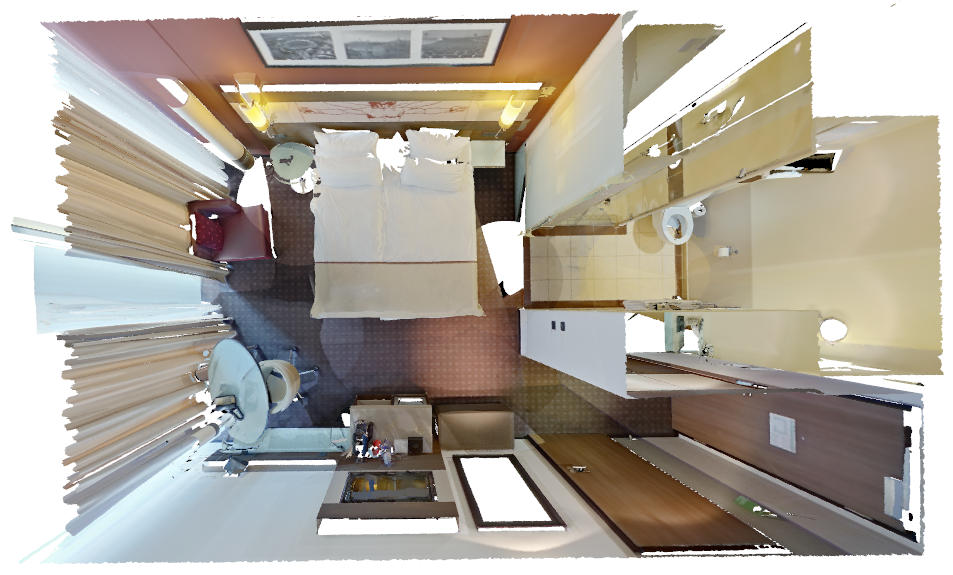} \\
    \includegraphics[width=\linewidth]{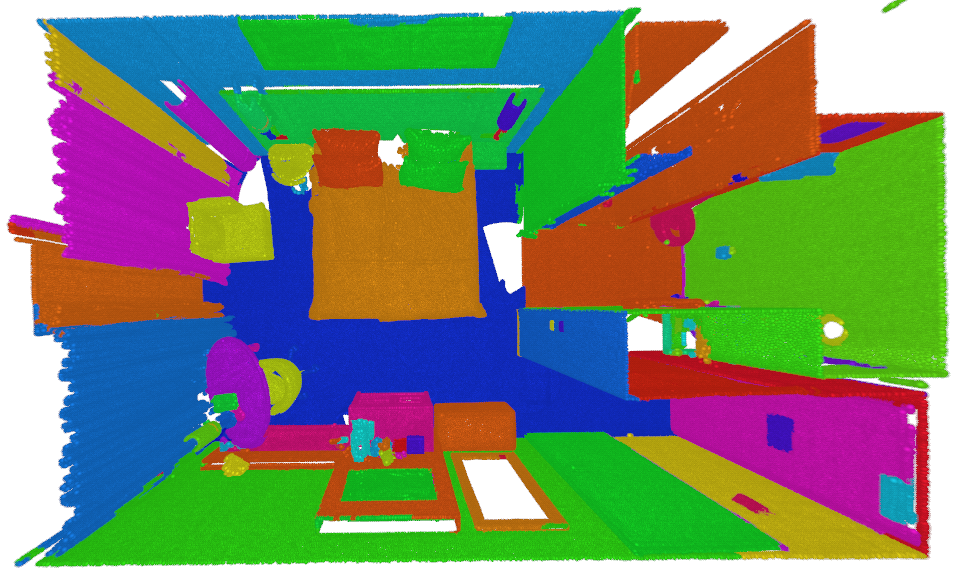} \\[-5pt]
    \caption*{6115eddb86}
  \end{minipage}
  \hfill
\begin{minipage}[b]{0.2\textwidth}
    \centering
    \includegraphics[width=\linewidth]{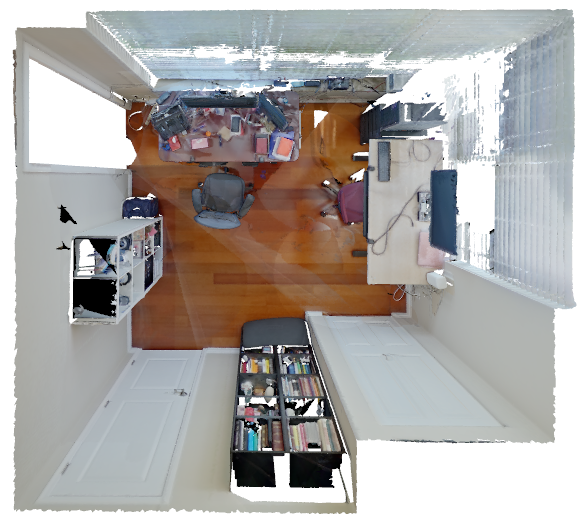} \\
    \includegraphics[width=\linewidth]{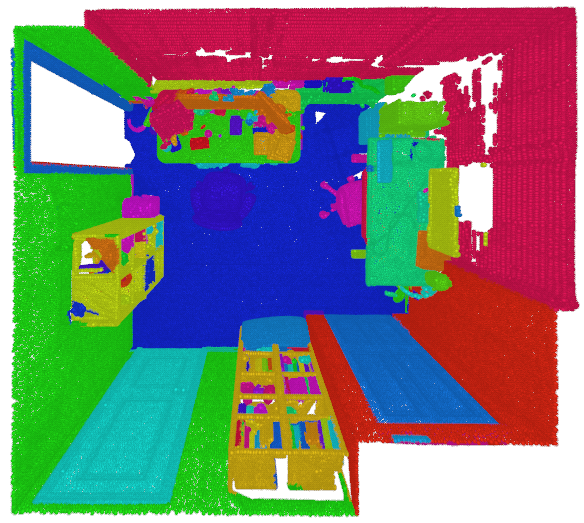} \\[-5pt]
    \caption*{cc5237fd77}
  \end{minipage}

  \caption{Qualitative Results on four scenes of the ScanNet++ validation split. Top row showing the mesh of the scan in a top-down view, the bottom row the respective point cloud with instance segmentation labels. The corresponding scene ID is shown below each segmentation.}

   \label{fig:qualitative_scannetpp}
\end{figure*}

To assess the performance of our method in the 3D space, we evaluate on the validation split of the ScanNet++ (v1) dataset.
Gaussians are initialized using a sampled scene point cloud containing up to 750k points.
For optimization, we uniformly select 300 frames per scene.
Due to the high resolution of the training images, optimization is performed at half resolution.
For inference on a scene, each point of the point cloud is assigned the instance ID of the nearest Gaussian mean.
We further apply post-processing using a graph-based smoothing method \cite{Felzenszwalb2004} commonly employed in related approaches \cite{schult2023mask3d,takmaz2023openmask3d,huang2025segment3d}.
Unlike Segment3D \cite{huang2025segment3d}, we omit the DBSCAN post-processing step, as it did not enhance results in our experiments. Each predicted instance is assigned to the semantic label with the highest similarity to the corresponding instance language embedding.

We report the official evaluation metrics, including the average precision $AP$ metric, computed over IoU thresholds ranging from $0.5$ to $0.95$ with a step size of $0.05$, and also the metrics $AP_{50}$ and $AP_{25}$ at IoU thresholds $0.5$ and $0.25$ respectively.

We first evaluate our method on the class-agnostic instance segmentation task following~\cite{huang2025segment3d}. The results detailed in Table~\ref{tab:scannetpp_val_agnostic} demonstrate that our approach significantly improves over the baseline methods. Additionally, the metrics reveal that our approach exhibits reduced dependency on post-processing as indicated by a smaller gap between the with and without post-processing variants.

Results on the ScanNet++ instance segmentation task on the validation split are shown in Table~\ref{tab:scannetpp_val_instance}. \ours operates in \emph{zero-shot open-vocabulary} setting and surpasses other open-vocabulary methods by a substantial margin (11.2 points over Segment3D on the $AP_{50}$ metric). Furthermore, \ours significantly narrows the performance gap to the \emph{fully-supervised closed-set} SGIFormer~\cite{yao2024sgiformer}. Our analysis indicates this mainly stems from the tendency of SAM to over-segment scenes. Therefore, addressing this issue might further reduce the gap.

Finally, we present qualitative results of the class-agnostic instance segmentation on four arbitrary ScanNet++ scenes in Figure~\ref{fig:qualitative_scannetpp}, showing the colored mesh in the top row from which the point clouds are sampled and our instance segmentation of the point cloud in the bottom row.

\begin{table}
    \centering
    \scriptsize
    \renewcommand{\tabcolsep}{3pt}
    \begin{tabularx}{\linewidth}{lYYYcYYY}
    \toprule
    \multirow{2}[2]{*}{Method} & \multicolumn{3}{c}{without post-processing} && \multicolumn{3}{c}{with post-processing} \\
    \cmidrule{2-4}
    \cmidrule{6-8}
    & AP & AP50 & AP25 && AP & AP50 & AP25 \\
    \midrule
    SAM3D \cite{yang2023sam3d} & ~~3.9 & ~~9.3 & 22.1 && ~~8.4 & 16.1 & $30.0$ \\
    Segment3D~\cite{huang2025segment3d} & 13.0 & 23.8 & 38.3 && 20.2 & $30.9$ & 42.7 \\
    \textit{Open3DIS}~\cite{nguyen2024open3dis}* & - & - & - && \underline{20.7} & \underline{38.6} & \underline{47.1} \\
    \midrule
OpenSplat3D (Ours) & \textbf{19.2} & \textbf{37.3} & \textbf{56.2} && \textbf{24.5} & \textbf{41.7} & \textbf{57.1} \\
\bottomrule
    \end{tabularx}
\vspace{-5pt}
    \caption{Class-agnostic instance segmentation on ScanNet++ val split. *: Open3DIS uses superpoints produced by Felzenszwalb and Huttenlocher segmentation~\cite{Felzenszwalb2004} directly in their pipeline.}
    \label{tab:scannetpp_val_agnostic}
\end{table}

\begin{table}
    \centering
    \scriptsize
    \renewcommand{\tabcolsep}{2pt}
    \begin{tabularx}{\linewidth}{lcccYYY}
    \toprule
    Method && Setting && AP & AP50 & AP25 \\
    \midrule
\textcolor{gray}{SGIFormer~\cite{yao2024sgiformer}} && \textcolor{gray}{fully-supervised} && \textcolor{gray}{23.9} & \textcolor{gray}{37.5} & \textcolor{gray}{46.6} \\
    \midrule
    Mask3D~\cite{schult2023mask3d} (+ OpenMask3D~\cite{takmaz2023openmask3d}) && open-vocabulary && - & 15.0 & - \\
    Segment3D~\cite{huang2025segment3d} (+ OpenMask3D~\cite{takmaz2023openmask3d}) && open-vocabulary && - & 18.5 & - \\
    \midrule
OpenSplat3D (Ours) && open-vocabulary && 16.5 & \textbf{29.7} & 39.0 \\
\bottomrule
    \end{tabularx}
    \vspace{-5pt}
    \caption{Instance Segmentation on the ScanNet++ validation split. Our method not only outperforms the other open-vocabulary methods by a large margin, it also reduces the gap to the state-of-the-art fully-supervised SGIFormer~\cite{yao2024sgiformer} approach.}
    \label{tab:scannetpp_val_instance}
\end{table}

\begin{figure}
    \centering
\begin{tikzpicture}
    \begin{axis}[
        xlabel={Variance loss weight $\lambda_\text{var}$},
        ylabel={mIoU},
        xmin=-0.1, xmax=4.1,
        ymin=0.65, ymax=0.85,
        xtick={0,1,2,3,4},
        ytick={0.65,0.7,0.75,0.8,0.85},
        legend pos=south east,
        grid=both,
        minor grid style={gray!25},
        major grid style={gray!25},
        width=\linewidth,
        height=3.5cm,
        font=\small,
        tick label style={font=\scriptsize}
    ]
    
    \addplot[
        color=blue,
        mark=*,
        line width=1pt,
        ]
        coordinates {
        (0.0,0.6759857277283379)
        (0.01,0.7814718508956394)
        (0.1,0.8169343146722433)
        (0.5,0.8306600802997255)
        (1.0,0.8335798779567877)
        (2.0,0.8333804988315144)
        (3.0,0.8322899111123215)
        (4.0,0.8346213202021214)
        };
    \end{axis}
    \end{tikzpicture}
    \vskip-2ex
    \caption{Influence of the variance loss, evaluated on the LERF-mask dataset.
    Even small $\lambda_\text{var}$ improves results significantly.}
    \label{fig:abl_var_grounded}
    \vspace{-1pt}
\end{figure}
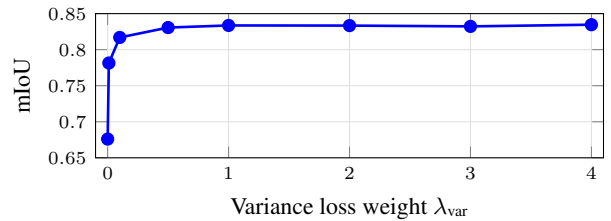

\PAR{Variance Loss Ablation.}
We conduct an ablation study on the LERF-mask dataset to analyze the effectiveness of our proposed variance regularization.
\cref{fig:abl_var_grounded} shows that disabling the variance loss leads to a substantial decrease in performance, and even a small weighting factor $\lambda_\text{var}$ significantly improves segmentation accuracy and this improvement is robust to the exact value.
These results strongly support our hypothesis that substantial feature blending occurs during the rendering process, and emphasize the importance of variance regularization for 3D instance learning by preserving the 2D to 3D consistency of the features of all Gaussians that affect a pixel.

\section{Conclusion}
\label{sec:conclusion}
\vspace{-3pt}

We propose OpenSplat3D, a method for open-vocabulary 3D instance segmentation, extending 3DGS with instance features and per-instance language embeddings.
Our novel variance regularization loss effectively improves feature consistency during alpha-composited rendering,
significantly improving the segmentation performance, while being efficient to compute.
Comprehensive evaluations across multiple open-vocabulary benchmarks demonstrate the effectiveness of OpenSplat3D, suggesting that 3D Gaussian representations in combination with 2D VFMs are an effective setup.
Future research directions may explore enhancements to instance learning, and different ways of incorporating language embeddings, but also extending the framework to dynamic environments is a promising direction.

{
\PAR{Acknowledgments.} This work was funded by Bosch Research as part of a Bosch-RWTH collaboration on ``Context Understanding for Autonomous Systems''.
}
\newpage
\clearpage
{
    \small
    \bibliographystyle{ieeenat_fullname}
    \bibliography{main}

\begin{thebibliography}{35}
\providecommand{\natexlab}[1]{#1}
\providecommand{\url}[1]{\texttt{#1}}
\expandafter\ifx\csname urlstyle\endcsname\relax
  \providecommand{\doi}[1]{doi: #1}\else
  \providecommand{\doi}{doi: \begingroup \urlstyle{rm}\Url}\fi

\bibitem[Campello et~al.(2013)Campello, Moulavi, and Sander]{hdbscan2013}
Ricardo J. G.~B. Campello, Davoud Moulavi, and Joerg Sander.
\newblock Density-based clustering based on hierarchical density estimates.
\newblock In \emph{Advances in Knowledge Discovery and Data Mining}, 2013.

\bibitem[Caron et~al.(2021)Caron, Touvron, Misra, J{\'e}gou, Mairal,
  Bojanowski, and Joulin]{caron2021dino}
Mathilde Caron, Hugo Touvron, Ishan Misra, Herv{\'e} J{\'e}gou, Julien Mairal,
  Piotr Bojanowski, and Armand Joulin.
\newblock Emerging properties in self-supervised vision transformers.
\newblock In \emph{ICCV}, 2021.

\bibitem[Choi et~al.(2024)Choi, Song, Kim, Kim, and Do]{choi2024click}
Seokhun Choi, Hyeonseop Song, Jaechul Kim, Taehyeong Kim, and Hoseok Do.
\newblock Click-{G}aussian: Interactive segmentation to any 3{D} gaussians.
\newblock In \emph{ECCV}, 2024.

\bibitem[Dai et~al.(2017)Dai, Chang, Savva, Halber, Funkhouser, and
  Nie{\ss}ner]{dai2017scannet}
Angela Dai, Angel~X. Chang, Manolis Savva, Maciej Halber, Thomas Funkhouser,
  and Matthias Nie{\ss}ner.
\newblock Scan{N}et: Richly-annotated 3{D} reconstructions of indoor scenes.
\newblock In \emph{CVPR}, 2017.

\bibitem[Engelmann et~al.(2024)Engelmann, Manhardt, Niemeyer, Tateno,
  Pollefeys, and Tombari]{engelmann2024opennerf}
Francis Engelmann, Fabian Manhardt, Michael Niemeyer, Keisuke Tateno, Marc
  Pollefeys, and Federico Tombari.
\newblock Open{NeRF}: Open set 3{D} neural scene segmentation with pixel-wise
  features and rendered novel views.
\newblock In \emph{ICLR}, 2024.

\bibitem[Felzenszwalb and Huttenlocher(2004)]{Felzenszwalb2004}
Pedro~F Felzenszwalb and Daniel~P Huttenlocher.
\newblock Efficient graph-based image segmentation.
\newblock \emph{IJCV}, 59, 2004.

\bibitem[Gu et~al.(2024)Gu, Lv, Frost, Green, Straub, and
  Sweeney]{gu2024egolifter}
Qiao Gu, Zhaoyang Lv, Duncan Frost, Simon Green, Julian Straub, and Chris
  Sweeney.
\newblock Ego{L}ifter: Open-world 3{D} segmentation for egocentric perception.
\newblock In \emph{ECCV}, 2024.

\bibitem[Han et~al.(2020)Han, Zheng, Xu, and Fang]{Han_2020_CVPR_Voxel}
Lei Han, Tian Zheng, Lan Xu, and Lu Fang.
\newblock Occu{S}eg: Occupancy-aware 3{D} instance segmentation.
\newblock In \emph{CVPR}, 2020.

\bibitem[Huang et~al.(2025)Huang, Peng, Takmaz, Tombari, Pollefeys, Song,
  Huang, and Engelmann]{huang2025segment3d}
Rui Huang, Songyou Peng, Ayca Takmaz, Federico Tombari, Marc Pollefeys, Shiji
  Song, Gao Huang, and Francis Engelmann.
\newblock Segment3{D}: Learning fine-grained class-agnostic 3{D} segmentation
  without manual labels.
\newblock In \emph{ECCV}, 2025.

\bibitem[Kerbl et~al.(2023)Kerbl, Kopanas, Leimk{\"u}hler, and
  Drettakis]{kerbl2023splatting}
Bernhard Kerbl, Georgios Kopanas, Thomas Leimk{\"u}hler, and George Drettakis.
\newblock {3D Gaussian Splatting for Real-Time Radiance Field Rendering}.
\newblock \emph{ACM TOG}, 2023.

\bibitem[Kerr et~al.(2023)Kerr, Kim, Goldberg, Kanazawa, and
  Tancik]{kerr2023lerf}
Justin Kerr, Chung~Min Kim, Ken Goldberg, Angjoo Kanazawa, and Matthew Tancik.
\newblock {LERF}: Language embedded radiance fields.
\newblock In \emph{CVPR}, 2023.

\bibitem[Kirillov et~al.(2023)Kirillov, Mintun, Ravi, Mao, Rolland, Gustafson,
  Xiao, Whitehead, Berg, Lo, et~al.]{kirillov2023sam}
Alexander Kirillov, Eric Mintun, Nikhila Ravi, Hanzi Mao, Chloe Rolland, Laura
  Gustafson, Tete Xiao, Spencer Whitehead, Alexander~C Berg, Wan-Yen Lo, et~al.
\newblock Segment anything.
\newblock In \emph{CVPR}, 2023.

\bibitem[Kolodiazhnyi et~al.(2024{\natexlab{a}})Kolodiazhnyi, Vorontsova,
  Konushin, and Rukhovich]{Kolodiazhnyi_2024_WACV_TopDownVoxel}
Maksim Kolodiazhnyi, Anna Vorontsova, Anton Konushin, and Danila Rukhovich.
\newblock Top-down beats bottom-up in 3{D} instance segmentation.
\newblock In \emph{WACV}, 2024{\natexlab{a}}.

\bibitem[Kolodiazhnyi et~al.(2024{\natexlab{b}})Kolodiazhnyi, Vorontsova,
  Konushin, and Rukhovich]{kolodiazhnyi2024oneformer3d}
Maxim Kolodiazhnyi, Anna Vorontsova, Anton Konushin, and Danila Rukhovich.
\newblock {OneFormer3D}: One transformer for unified point cloud segmentation.
\newblock In \emph{CVPR}, 2024{\natexlab{b}}.

\bibitem[Liu et~al.(2024)Liu, Zeng, Ren, Li, Zhang, Yang, Jiang, Li, Yang, Su,
  et~al.]{liu2025grounding}
Shilong Liu, Zhaoyang Zeng, Tianhe Ren, Feng Li, Hao Zhang, Jie Yang, Qing
  Jiang, Chunyuan Li, Jianwei Yang, Hang Su, et~al.
\newblock Grounding {DINO}: Marrying {DINO} with grounded pre-training for
  open-set object detection.
\newblock In \emph{ECCV}, 2024.

\bibitem[Mildenhall et~al.(2020)Mildenhall, Srinivasan, Tancik, Barron,
  Ramamoorthi, and Ng]{mildenhall2020nerf}
Ben Mildenhall, Pratul~P. Srinivasan, Matthew Tancik, Jonathan~T. Barron, Ravi
  Ramamoorthi, and Ren Ng.
\newblock {NeRF}: Representing scenes as neural radiance fields for view
  synthesis.
\newblock In \emph{ECCV}, 2020.

\bibitem[Nguyen et~al.(2024)Nguyen, Ngo, Kalogerakis, Gan, Tran, Pham, and
  Nguyen]{nguyen2024open3dis}
Phuc Nguyen, Tuan~Duc Ngo, Evangelos Kalogerakis, Chuang Gan, Anh Tran, Cuong
  Pham, and Khoi Nguyen.
\newblock Open3{DIS}: Open-vocabulary 3{D} instance segmentation with 2{D} mask
  guidance.
\newblock In \emph{CVPR}, 2024.

\bibitem[Oquab et~al.(2024)Oquab, Darcet, Moutakanni, Vo, Szafraniec, Khalidov,
  Fernandez, Haziza, Massa, El-Nouby, et~al.]{oquab2023dinov2}
Maxime Oquab, Timoth{\'e}e Darcet, Th{\'e}o Moutakanni, Huy Vo, Marc
  Szafraniec, Vasil Khalidov, Pierre Fernandez, Daniel Haziza, Francisco Massa,
  Alaaeldin El-Nouby, et~al.
\newblock {DINOv2}: Learning robust visual features without supervision.
\newblock \emph{TMLR}, 2024.

\bibitem[Peng et~al.(2023)Peng, Genova, Jiang, Tagliasacchi, Pollefeys,
  Funkhouser, et~al.]{peng2023openscene}
Songyou Peng, Kyle Genova, Chiyu Jiang, Andrea Tagliasacchi, Marc Pollefeys,
  Thomas Funkhouser, et~al.
\newblock Open{S}cene: 3{D} scene understanding with open vocabularies.
\newblock In \emph{CVPR}, 2023.

\bibitem[Qin et~al.(2024)Qin, Li, Zhou, Wang, and Pfister]{qin2024langsplat}
Minghan Qin, Wanhua Li, Jiawei Zhou, Haoqian Wang, and Hanspeter Pfister.
\newblock Lang{S}plat: 3{D} language gaussian splatting.
\newblock In \emph{CVPR}, 2024.

\bibitem[Radford et~al.(2021)Radford, Kim, Hallacy, Ramesh, Goh, Agarwal,
  Sastry, Askell, Mishkin, Clark, et~al.]{radford2021clip}
Alec Radford, Jong~Wook Kim, Chris Hallacy, Aditya Ramesh, Gabriel Goh,
  Sandhini Agarwal, Girish Sastry, Amanda Askell, Pamela Mishkin, Jack Clark,
  et~al.
\newblock Learning transferable visual models from natural language
  supervision.
\newblock In \emph{ICML}, 2021.

\bibitem[Raschka et~al.(2020)Raschka, Patterson, and Nolet]{raschka2020cuml}
Sebastian Raschka, Joshua Patterson, and Corey Nolet.
\newblock Machine learning in {P}ython: Main developments and technology trends
  in data science, machine learning, and artificial intelligence.
\newblock \emph{Information}, 11\penalty0 (4):\penalty0 193, 2020.

\bibitem[Schult et~al.(2023)Schult, Engelmann, Hermans, Litany, Tang, and
  Leibe]{schult2023mask3d}
Jonas Schult, Francis Engelmann, Alexander Hermans, Or Litany, Siyu Tang, and
  Bastian Leibe.
\newblock Mask3{D}: Mask transformer for 3{D} semantic instance segmentation.
\newblock In \emph{ICRA}, 2023.

\bibitem[Shi et~al.(2024)Shi, Wang, Duan, and Guan]{shi2024legaussian}
Jin-Chuan Shi, Miao Wang, Hao-Bin Duan, and Shao-Hua Guan.
\newblock Language embedded 3{D} gaussians for open-vocabulary scene
  understanding.
\newblock In \emph{CVPR}, 2024.

\bibitem[Silva et~al.(2024)Silva, Dahaghin, Toso, and
  Del~Bue]{silva2024contrastive}
Myrna~C. Silva, Mahtab Dahaghin, Matteo Toso, and Alessio Del~Bue.
\newblock Contrastive gaussian clustering for weakly supervised 3{D} scene
  segmentation.
\newblock In \emph{ICPR}, 2024.

\bibitem[Sun et~al.(2023)Sun, Qing, Tan, and Xu]{SunSPFormer2023}
Jiahao Sun, Chunmei Qing, Junpeng Tan, and Xiangmin Xu.
\newblock Superpoint transformer for 3{D} scene instance segmentation.
\newblock \emph{AAAI}, 2023.

\bibitem[Takmaz et~al.(2023)Takmaz, Fedele, Sumner, Pollefeys, Tombari, and
  Engelmann]{takmaz2023openmask3d}
Ay{\c{c}}a Takmaz, Elisabetta Fedele, Robert~W. Sumner, Marc Pollefeys,
  Federico Tombari, and Francis Engelmann.
\newblock {OpenMask3D: Open-Vocabulary 3D Instance Segmentation}.
\newblock In \emph{NeurIPS}, 2023.

\bibitem[Wang et~al.(2004)Wang, Bovik, Sheikh, and Simoncelli]{ssim2004}
Zhou Wang, A.C. Bovik, H.R. Sheikh, and E.P. Simoncelli.
\newblock Image quality assessment: from error visibility to structural
  similarity.
\newblock \emph{TIP}, 13\penalty0 (4):\penalty0 600--612, 2004.

\bibitem[Wu et~al.(2024)Wu, Meng, Li, Wu, Shi, Cheng, Zhao, Feng, Ding, Wang,
  and Zhang]{wu2024opengaussian}
Yanmin Wu, Jiarui Meng, Haijie Li, Chenming Wu, Yahao Shi, Xinhua Cheng, Chen
  Zhao, Haocheng Feng, Errui Ding, Jingdong Wang, and Jian Zhang.
\newblock Open{G}aussian: Towards point-level 3{D} gaussian-based open
  vocabulary understanding.
\newblock In \emph{NeurIPS}, 2024.

\bibitem[Xu et~al.(2023)Xu, Xiong, Ding, and Tu]{xu2023masqclip}
Xin Xu, Tianyi Xiong, Zheng Ding, and Zhuowen Tu.
\newblock Mas{QCLIP} for open-vocabulary universal image segmentation.
\newblock In \emph{ICCV}, 2023.

\bibitem[Yang et~al.(2023)Yang, Wu, He, Zhao, and Liu]{yang2023sam3d}
Yunhan Yang, Xiaoyang Wu, Tong He, Hengshuang Zhao, and Xihui Liu.
\newblock {SAM3D}: Segment anything in 3{D} scenes.
\newblock In \emph{ICCV Workshops}, 2023.

\bibitem[Yao et~al.(2024)Yao, Wang, Liu, and Chau]{yao2024sgiformer}
Lei Yao, Yi Wang, Moyun Liu, and Lap-Pui Chau.
\newblock {SGIF}ormer: Semantic-guided and geometric-enhanced interleaving
  transformer for 3{D} instance segmentation.
\newblock \emph{TCSVT}, 2024.

\bibitem[Ye et~al.(2024)Ye, Danelljan, Yu, and Ke]{ye2024gaussiangrouping}
Mingqiao Ye, Martin Danelljan, Fisher Yu, and Lei Ke.
\newblock Gaussian {G}rouping: Segment and edit anything in 3{D} scenes.
\newblock In \emph{ECCV}, 2024.

\bibitem[Yeshwanth et~al.(2023)Yeshwanth, Liu, Nie{\ss}ner, and
  Dai]{yeshwanth2023scannet++}
Chandan Yeshwanth, Yueh-Cheng Liu, Matthias Nie{\ss}ner, and Angela Dai.
\newblock Scan{N}et++: A high-fidelity dataset of 3{D} indoor scenes.
\newblock In \emph{CVPR}, 2023.

\bibitem[Zhai et~al.(2023)Zhai, Mustafa, Kolesnikov, and Beyer]{zhai2023siglip}
Xiaohua Zhai, Basil Mustafa, Alexander Kolesnikov, and Lucas Beyer.
\newblock Sigmoid loss for language image pre-training.
\newblock In \emph{ICCV}, 2023.

\end{thebibliography}
}

\end{document}